\documentclass[runningheads]{llncs}
\usepackage{graphicx}
\usepackage{grffile}
\usepackage{comment}
\usepackage{amsmath,amssymb}
\usepackage{color}
\usepackage{multirow}
\usepackage{pdfpages}
\usepackage{textcomp}

\usepackage[inline]{enumitem}
\usepackage{subcaption}
\usepackage{bm}
\usepackage{xspace}
\usepackage{booktabs}

\newcommand{\modelname}{\mbox{ExPose}\xspace}
\newcommand{\modelnameExplain}{{EXpressive POse and Shape rEgression}\xspace}

\newcommand{\websiteURL}{\mbox{\url{https://expose.is.tue.mpg.de}}}

\newcommand{\ourTitle}{Monocular Expressive Body Regression \\through Body-Driven Attention}

\newcommand{\highlightCR}[1]{{\textcolor{black}{{#1}}}}
\newcommand{\na}{\mbox{N/A}\xspace}

\usepackage{pifont}
\newcommand{\cmark}{\color{green}\ding{51}}
\newcommand{\xmark}{\color{red}\ding{55}}

\newcommand{\spin}{\mbox{SPIN}\xspace}
\newcommand{\smplx}{\mbox{SMPL-X}\xspace}
\newcommand{\smplifyx}{\mbox{SMPLify-X}\xspace}
\newcommand{\groundtruth}{\mbox{ground-truth}\xspace}
\newcommand{\pseudogt}{pseudo \mbox{ground-truth}\xspace}
\newcommand{\stateoftheart}{\mbox{state-of-the-art}\xspace}
\newcommand{\freihand}{\mbox{FreiHAND}\xspace}

\newcommand{\fpn}{\mbox{FPN}\xspace}
\newcommand{\detectron}{\mbox{Detectron2}\xspace}
\newcommand{\adam}{\mbox{Adam}\xspace}
\newcommand{\ghum}{\mbox{GHUM}\xspace}
\newcommand{\ghuml}{\mbox{GHUML}\xspace}
\newcommand{\smpl}{\mbox{SMPL}\xspace}

\newcommand{\mano}{\mbox{MANO}\xspace}
\newcommand{\flame}{\mbox{FLAME}\xspace}

\newcommand{\resnet}{\mbox{ResNet50}\xspace}
\newcommand{\ringnet}{\mbox{RingNet}\xspace}

\newcommand{\pytorch}{\mbox{PyTorch}\xspace}
\newcommand{\mocap}{\mbox{MoCap}\xspace}

\newcommand{\inthewild}{\mbox{in-the-wild}\xspace}
\newcommand{\twoD}{2D\xspace}
\newcommand{\threeD}{3D\xspace}

\newcommand{\supmat}{Sup.~Mat.\xspace}

\newcommand{\ehf}{\mbox{EHF}\xspace}

\newcommand{\mtc}{\mbox{MTC}\xspace}

\newcommand{\mesh}{M}
\newcommand{\shape}{\bm{\beta}}
\newcommand{\pose}{\bm{\theta}}
\newcommand{\expression}{\bm{\psi}}

\newcommand{\face}{f}

\newcommand{\labelLEFT}{\mbox{\emph{Left:}}\xspace}
\newcommand{\labelMIDDLE}{\mbox{\emph{Middle:}}\xspace}
\newcommand{\labelRIGHT}{\mbox{\emph{Right:}}\xspace}

\newcommand{\etal}{et al.\xspace}
\newcommand{\ie}{i.e.\xspace}
\newcommand{\eg}{e.g.\xspace}

\newcommand{\normabs}[1]{\left\lVert#1\right\rVert_1}
\newcommand{\normmse}[1]{\left\lVert#1\right\rVert_2^2}

\newcommand{\rgb}{\mbox{RGB}\xspace}
\newcommand{\rgbD}{\mbox{RGB-D}\xspace}

\newcommand{\highlight}[1]{\xspace{\color{black} #1}\xspace}
\newcommand{\labelBODY}[1]{\xspace{\color{black} \textbf{\emph{#1}}}\xspace}

\begin{document}
\pagestyle{headings}
\mainmatter

\title{\ourTitle}
\titlerunning{\ourTitle}
\author{
Vasileios Choutas\inst{1,2}	\and
Georgios Pavlakos\inst{3}	\and
Timo Bolkart	\inst{1}			\and\\
Dimitrios Tzionas\inst{1}	\and
Michael J.~Black\inst{1}
}

\authorrunning{V. Choutas \etal} %

\institute{
\scriptsize
$^1$ Max Planck Institute for Intelligent Systems, T{\"u}bingen, Germany	\\
$^2$ Max Planck ETH Center for Learning Systems							\\
$^3$ University of Pennsylvania, Philadelphia, USA 						\\
\email{\{vchoutas, gpavlakos, tbolkart, dtzionas, black\}@tuebingen.mpg.de}
}

\maketitle

\begin{abstract}
To understand how people look, interact, or perform tasks, we need to quickly and accurately capture their \threeD body, face, and hands \emph{together} from an \rgb image.
Most existing methods focus only on parts of the body. %
A few recent approaches reconstruct full expressive \threeD humans from images
using \threeD body models that include the face and hands.
These methods are optimization-based and thus slow, prone to local
optima, and require \twoD keypoints as input.
We address these limitations by introducing \emph{\modelname} (\modelnameExplain), which
directly regresses the body, face, and hands, in \smplx  format, from an \rgb image.
This is a hard problem due to the high dimensionality of the body and the lack
of expressive training data.
Additionally, hands and faces are much smaller than the body,
occupying very few image pixels.
This makes hand and face estimation hard when body images are downscaled for neural networks.
We make three main contributions.
First, we account for the lack of training data by curating a \emph{dataset} of \smplx fits on \inthewild images. %
Second, we observe that body estimation localizes the face and hands reasonably well.
We introduce \emph{body-driven attention} for face and hand regions in
the original image to extract higher-resolution crops that are fed to dedicated refinement modules.
Third, these modules exploit \emph{part-specific knowledge} from
existing face- and hand-only datasets. %
\modelname estimates expressive \threeD humans more accurately than
existing optimization methods at a small fraction of the computational cost.
Our data, model and code are available for research at \websiteURL.
\end{abstract}

\section{Introduction}
\label{sec:introduction}

\begin{figure}[t]
    \centering
    \centering      	   %
    \includegraphics[trim=000mm 000mm 000mm 000mm, clip=false, height=0.335\textwidth]{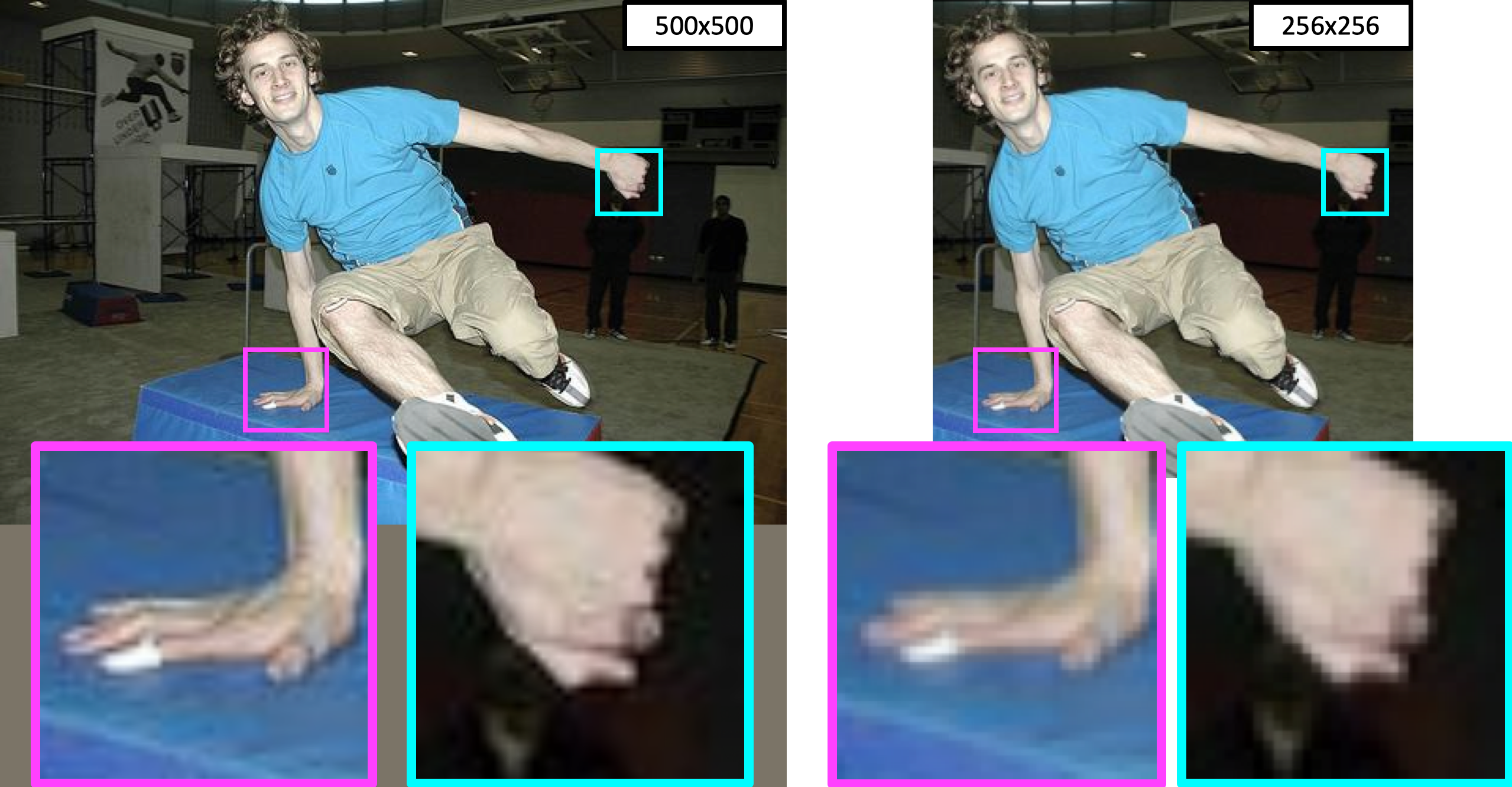}
    \includegraphics[trim=000mm 000mm 000mm 000mm, clip=false, height=0.335\textwidth]{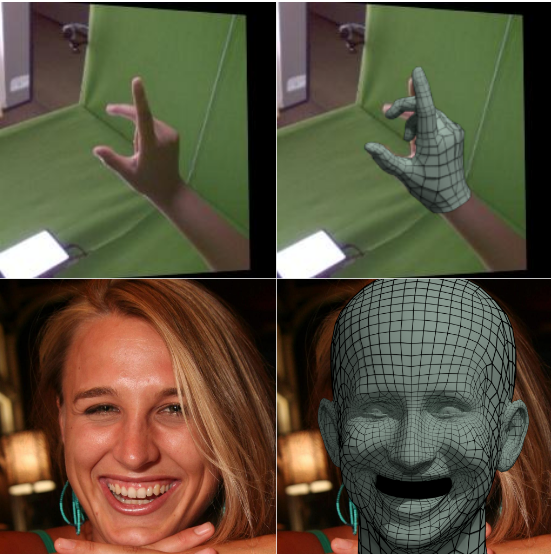}
    \caption{
        \labelLEFT		Full-body RGB images of people contain
        many more pixels on the body than on the face or hands.
        \labelMIDDLE	Images are typically downsized (\eg to
        \mbox{$256 \times 256$ px}) for use in neural networks.
        This resolution is fine for the body but the hands and face
        suffer from low resolution. %
        Our model (Figure~\ref{fig:arch})
        uses \emph{body-driven attention} to restore the lost information for hands and faces from the original image, feeding it to dedicated refinement modules.
        \labelRIGHT		These modules give more expressive hands and faces, by exploiting \emph{part-specific knowledge}
        learned from higher quality hand-only~\cite{Freihand2019} and
        face-only~\cite{karras2019style} datasets; green meshes show
        example part-specific training
        data.
    }
    \label{fig:problem_DownScale}
\end{figure}

A long term goal of computer vision is to understand humans and their behavior in everyday scenarios using only images.
Are they happy or sad?  How do they interact with each other and the physical world?   What are their intentions?
To answer such difficult questions, we first need to \emph{quickly} and \emph{accurately} reconstruct their \threeD body, face and hands \emph{together} from a single RGB image.
This is very challenging.
As a result, the community has broken the problem into pieces with much of the work
focused on estimating either the
main body 	\cite{Review_Gavrila,Review_Moeslund_2006,Sarafianos:Survey:2016},
the face 			\cite{zollhofer2018state}
or the hands 		\cite{Review_Erol_HandPose,Supancic:ICCV:2015,yuan2018depth}
separately.

Only recent advances have made the problem tractable in its full complexity. %
Early methods estimate \twoD joints and features \cite{cao2018openpose,Hidalgo_2019_ICCV} for the body, face and hands.
However, this is not enough.
It is the skin surface that describes important aspects of humans, \eg what their precise \threeD shape is, whether they are smiling, gesturing or holding something.
For this reason, strong statistical parametric models for expressive \threeD humans were introduced, namely \adam \cite{joo2018total}, \smplx \cite{Pavlakos_2019_CVPR} and recently
\ghum/\ghuml~\cite{Xu_2020_CVPR}.
Such models are attractive because they facilitate reconstruction even from ambiguous data, working as a strong prior.

There exist three methods that estimate full expressive \threeD humans from an
\rgb image \cite{Pavlakos_2019_CVPR,Xiang_2019_CVPR,Xu_2020_CVPR}, using \smplx, \adam and \ghum/\ghuml respectively.
These methods are based on optimization, therefore they are slow, prone to local optima, and rely on heuristics for initialization.
These issues significantly limit the applicability of these methods.
In contrast, recent body-only methods \cite{kanazawa_cvpr_2018,Kolotouros_2019_ICCV} directly regress \threeD \smpl bodies quickly and relatively reliably from an \rgb image.

Here we present a \emph{fast} and \emph{accurate} model that reconstructs full \emph{expressive} \threeD humans, by estimating \smplx parameters directly from an \rgb image. %
This is a hard problem and we show that it is not easily solved by extending \smpl neural-network regressors to \smplx for several reasons.
First, \smplx is a much higher dimensional model than \smpl.
Second, there exists no large \inthewild dataset with \smplx annotations for training.
Third, the face and hands are often blurry and occluded in images.
At any given image resolution, they also occupy many fewer pixels than the body, making them low resolution.
Fourth, for technical reasons, full body images are typically downscaled for input to neural networks \cite{krizhevsky2012imagenet}, \eg to $256 \times 256$ pixels.
As shown in Figure~\ref{fig:problem_DownScale}, this results in even lower resolution for the hands and face, making inference difficult.

Our model and training method, shown in Figure~\ref{fig:arch}, tackles all these challenges.
We account for data scarcity by introducing a new dataset with paired \inthewild images and \smplx annotations.
To this end, we employ several standard \inthewild body datasets \cite{andriluka20142d,johnson2010clustered,johnson2011learning,lin2014microsoft} and fit \smplx to them with \smplifyx~\cite{Pavlakos_2019_CVPR}.
We semi-automatically curate these fits to keep only the good ones as \pseudogt.
We then train a model that regresses \smplx parameters from an \rgb image, similar to \cite{kanazawa_cvpr_2018}.
However, this only estimates rough hand and face configurations, due to the problems described above.
We observe that the main body is estimated well, on par with \cite{kanazawa_cvpr_2018,Kolotouros_2019_ICCV}, providing good rough
localization for the face and hands.
We use this for \emph{body-driven attention} and focus the network back on the \emph{original} \mbox{non-downscaled} image for the face and hands.
We retrieve high-resolution information for these regions and feed this to dedicated \emph{refinement} modules.
These modules act as an \emph{expressivity boost} by distilling \emph{\mbox{part-specific} knowledge} from high-quality hand-only \cite{Freihand2019} and face-only \cite{liu2015celebA} datasets.
Finally, the independent components are fine-tuned jointly end-to-end,
so that the part networks can benefit from the full-body initialization.

We call the final model \modelname (\modelnameExplain).
\modelname is at least as accurate as existing optimization-based methods \cite{Pavlakos_2019_CVPR} for estimating expressive \threeD humans, while running two orders of magnitude faster.
Our \highlight{data, model and code} are available for research at \websiteURL.

\section{Related Work}
\label{sec:related_work}

\textbf{Human Modeling:}
Modeling and capturing the whole human body is a challenging problem.
To make it tractable, the community has studied the body, face and hands separately, in a \mbox{divide-and-conquer} fashion.
For the human \labelBODY{face}, the seminal work of Blanz and Vetter \cite{blanzvetter1999} introduces the first \threeD morphable model.
Since then, numerous works (see \cite{egger20193d}) propose more powerful face models and methods to infer their parameters.
For human \labelBODY{hands} the number of models is limited, with
Khamis \etal \cite{msr_2015_cvpr_learnshapemodel} learning a model of hand shape variation from depth images, while
Romero \etal \cite{romero2017embodied} learn a parametric hand model with both a rich shape and pose space from \threeD hand scans.
For the human \labelBODY{body}, the introduction of the CAESAR dataset \cite{CAESAR} enables the creation of models that disentangle shape and pose,
such as SCAPE \cite{Anguelov05} and \smpl \cite{SMPL:2015}, to name a few.
However, these models have a neutral face  and the hands are non-articulated. %
 In contrast, \adam \cite{joo2018total} and \smplx
 \cite{Pavlakos_2019_CVPR} are the first models that represent the
 body, face and hands jointly.
\adam lacks the pose-dependent blendshapes of \smpl and the released
version does not include a face model.
The \ghum~\cite{Xu_2020_CVPR} model is similar to \smplx but is not publicly available at
the time of writing.

\textbf{Human Pose Estimation:}
Often pose estimation is posed as the estimation of \twoD or \threeD keypoints, corresponding to anatomical joints or landmarks \cite{bulat2017far,cao2018openpose,simon2017hand}.
In contrast, recent advances use richer representations of  the
\threeD  body surface in the form of parametric \cite{bogo2016keep,kanazawa_cvpr_2018,omran2018neural,pavlakos2018learning}
or non-parametric
\cite{kolotouros2019convolutional,Saito_2019_ICCV,varol2018bodynet}
models. %

To estimate \labelBODY{bodies} from images, many methods break the
problem down into stages.
First, they estimate some intermediate representation such as
\twoD joints \cite{bogo2016keep,grauman2003inferring,guan_iccv_scape_2009,MuVS_3DV_2017,kanazawa_cvpr_2018,martinez_2017_3dbaseline,pavlakos2018learning,sigal2006predicting,tome2017lifting,zhao_2019_cvpr_semgcn},
silhouettes \cite{agarwal_trigs_3d_poses,MuVS_3DV_2017,pavlakos2018learning}, part labels \cite{omran2018neural,Ruegg:AAAI:2020} or dense correspondences \cite{Guler_densepose,yu_iccv2019_hybrid}.
Then, they reconstruct the body pose out of this proxy information, by either using it in the data term of an optimized energy function \cite{bogo2016keep,MuVS_3DV_2017,zanfir_2018_cvpr} or
``lifting'' it using a trained regressor \cite{kanazawa_cvpr_2018,martinez_2017_3dbaseline,omran2018neural,pavlakos2018learning,tome2017lifting}.
Due to ambiguities in lifting \twoD to \threeD, such methods use various priors for regularization,
such as known limb lengths \cite{lee1985determination}, a pose prior for joint angle limits \cite{akhter2015pose},
or a statistical body model \cite{bogo2016keep,MuVS_3DV_2017,omran2018neural,pavlakos2018learning} like \smpl \cite{SMPL:2015}.
The above \twoD proxy representations have the advantage that
annotation for them is readily available.
Their disadvantage is that the eventual regressor does not get to exploit the
original image pixels and errors made by the proxy task cannot be overcome.

Other methods predict \threeD pose directly from \rgb pixels.
Intuitively, they have to learn a harder mapping, but they avoid information bottlenecks and additional sources of error.
Most methods infer \threeD body joints \cite{li2015maximum,pavlakos2017coarse,sun2017compositional,sun2018integral,bugra_bmvc_2016},
parametric methods estimate model parameters~\cite{kanazawa_cvpr_2018,kanazawa_2019_cvpr,Kolotouros_2019_ICCV},
while non-parametric methods estimate \threeD meshes
\cite{kolotouros2019convolutional},
depth maps~\cite{Gabeur_2019_ICCV,Smith_2019_ICCV}
voxels~\cite{varol2018bodynet,Zheng_2019_ICCV} or distance
fields~\cite{Saito_2019_ICCV,Saito_2020_CVPR}.
Datasets of paired indoor images and \mocap data~\cite{ionescupapavaetal2014,sigal_ijcv_10b} allow supervised training,
but may not generalize to \inthewild data.
To account for this, Rogez and Schmid \cite{rogez2016mocap} augment these datasets by overlaying synthetic \threeD humans, while
Kanazawa \etal~\cite{kanazawa_cvpr_2018} include \inthewild datasets~\cite{andriluka20142d,johnson2010clustered,johnson2011learning,lin2014microsoft}
and employ a \mbox{re-projection} loss on their \twoD joint annotations for weak supervision.

Similar observations can be made in the human hand and face literature.
For \labelBODY{hands}, there has been a lot of work on \rgbD data
\cite{yuan2018depth}, and more recent interest in
monocular \rgb \cite{Baek_2019_CVPR,boukhayma_2019_cvpr,honnotate2020,hasson_2019_cvpr,Iqbal_2018_ECCV,kulon2019rec,Mueller_2018_GANerated,Tekin_2019_CVPR,brox_ICCV_2017}.
Some of the non-parametric methods estimate \threeD joints \cite{Iqbal_2018_ECCV,Mueller_2018_GANerated,Tekin_2019_CVPR,brox_ICCV_2017},
while others estimate \threeD meshes \cite{Ge_2019_CVPR,Kulon_2020_CVPR}.
Parametric models \cite{Baek_2019_CVPR,boukhayma_2019_cvpr,hasson_2019_cvpr,kulon2019rec,Zhang_2019_ICCV}
estimate configurations of statistical models like \mano \cite{romero2017embodied}
or a graph morphable model \cite{kulon2019rec}.
For \labelBODY{faces}, \threeD reconstruction and tracking
has a long history. We refer the reader to a recent comprehensive
survey \cite{zollhofer2018state}.

\textbf{Attention for Human Pose Estimation:}
In the context of human pose estimation,
attention is often used to improve
prediction accuracy.
Successful architectures for 2D pose estimation,
like Convolutional Pose Machines~\cite{wei2016convolutional}
and Stacked Hourglass~\cite{newell2016stacked}
include a series of processing stages,
where the intermediate pose predictions in the form of heatmaps
are used as input to the following stages.
This informs the network of early predictions
and guides its attention to relevant image pixels.
Chu~\etal~\cite{chu2017multi} build explicit attention maps,
at a global and part-specific level,
driving the model to focus on regions of interest.
Instead of predicting attention maps,
our approach uses the initial body mesh prediction
to define the areas of attention for hands- and face-specific
processing networks.
A similar practice is used by OpenPose~\cite{cao2018openpose},
where arm keypoints are used to estimate hand bounding boxes,
in a heuristic manner.
Additionally, for HoloPose~\cite{Guler_2019_CVPR},
body keypoints are used to pool
part-specific features from the image.

A critical difference of \modelname is that,
instead of simply pooling already
computed features,
we also process the region of interest
at  higher resolution, to capture
more subtle face and hand details.
In related work, Chandran \etal~\cite{Chandran_2020_CVPR} use a low resolution proxy image to
detect facial landmarks and extract high resolution crops that are
used to refine facial landmark predictions.

\textbf{Expressive Human Estimation:}
Since expressive
parametric models of the human
body have only recently been introduced~\cite{joo2018total,Pavlakos_2019_CVPR,romero2017embodied,Xu_2020_CVPR},
there are only a few methods to reconstruct their parameters.
Joo~\etal~\cite{joo2018total} present an early approach,
but rely on an extended multi-view setup.
More recently, Xiang~\etal~\cite{Xiang_2019_CVPR},
Pavlakos~\etal~\cite{Pavlakos_2019_CVPR} and Xu~\etal~\cite{Xu_2020_CVPR}
use a single image to recover
\adam, \smplx and \ghum parameters respectively,
using optimization-based approaches.
This type of inference can be slow and
may fail in the presence of noisy feature detections.
In contrast, we present the first
regression approach for expressive monocular capture and show that it
is both more accurate and significantly faster than prior work.

\section{Method}
\label{sec:method}

\begin{figure}[t!]
    \centering      	   %
	\includegraphics[trim=000mm 000mm 000mm 000mm, clip=false, width=1.00\textwidth]{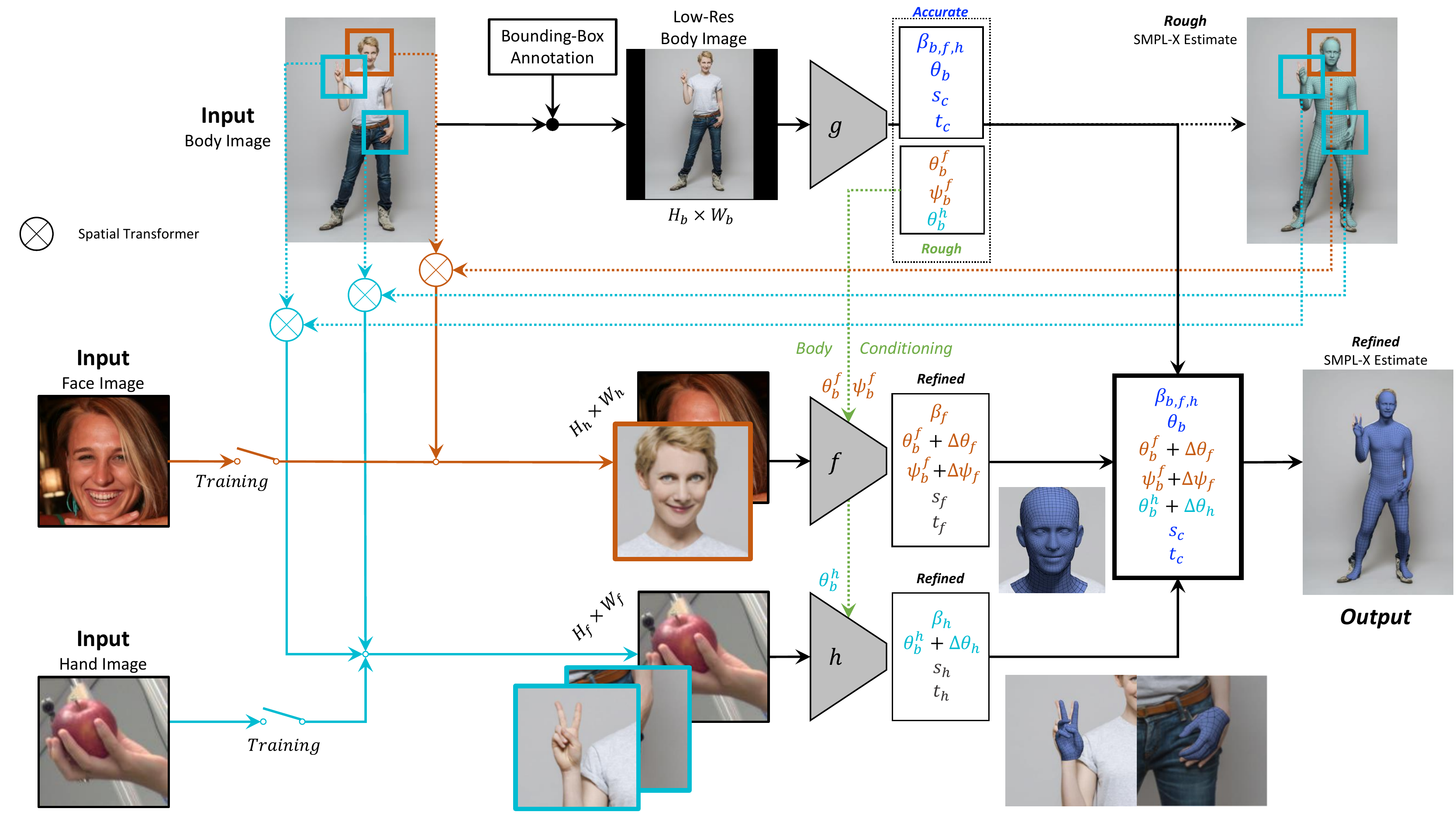}
        \caption{
            An image of the body is extracted using a bounding box from the full
            resolution image and fed to a neural network $g(\cdot)$, that predicts
            body pose $\pose_b$, hand
            pose $\pose_h$, facial pose $\pose_{\text{f}}$, shape
            $\shape$, expression $\expression$, camera scale $s$ and translation
            $\bm{t}$. Face and hand images are extracted from the original
            resolution image using bilinear interpolation. These are fed to part
            specific sub-networks $f(\cdot)$ and $h(\cdot)$ respectively to produce the
            final estimates for the face and hand parameters.
            During training the part specific networks can also receive hand and face only data
            for extra supervision.
        }
        \label{fig:arch}
    \end{figure}

\subsection{\threeD Body Representation}
\label{sec:technical_body_model}

To represent the human body, we use \smplx~\cite{Pavlakos_2019_CVPR},
a generative model that captures shape variation,
limb articulation and facial expressions across a human population. It is learned
from a collection of registered \threeD body, hand and face scans of people with
different sizes, nationalities and genders.
The shape, $\shape \in \mathbb{R}^{10}$, and expression, $\expression \in \mathbb{R}^{10}$,
are described by 10 coefficients from the corresponding PCA spaces.
The articulation of the limbs, the hands and the face is
modeled by the pose vector $\pose \in \mathbb{R}^{J\times D}$,
where $D$ is the rotation representation dimension, \eg $D=3$ if we select
axis-angles,
which describes the
relative rotations of the $J = 53$ major joints.
These joints include $22$ main body joints, $1$ for the jaw,
and $15$ joints per hand for the fingers.
\smplx is a differentiable function $\mesh(\shape, \pose, \expression)$,
that produces a \threeD mesh $\mesh = (V,
F)$ for the human body, with $N=10475$ vertices $V \in
\mathbb{R}^{(N \times 3)}$ and triangular faces $F$. %
The surface of the articulated body is obtained by linear blend skinning driven by a
rigged skeleton, defined by the above joints.
Following the notation of~\cite{kanazawa_cvpr_2018}
we denote posed joints with $X(\pose, \shape) \in \mathbb{R}^{J\times 3}$.
The final set of \smplx parameters is the vector $\Theta = \left\{\shape, \pose,
\expression\right\} \in \mathbb{R}^{338}$, as we choose to represent the pose
parameters $\pose$
using the representation of Zhou \etal~\cite{zhou_2019_cvpr} with $D=6$.

\subsection{Body-driven Attention}
\label{subsec:model_training}

Instead of attempting to regress body, hand and face parameters from a low resolution image crop
we design an attentive architecture that uses the structure of the body and the
full resolution of the image $I$.
Given a bounding box of the body,
we extract an image $I_b$, using an affine transformation $T_b \in
\mathbb{R}^{2 \times 3}$, from the high resolution image $I$.
The body crop $I_b$ is fed to a neural network $g$, similar to~\cite{kanazawa_cvpr_2018},
to produce a first set of \smplx parameters $\Theta_b$ and weak-perspective
camera scale $s_b \in \mathbb{R}$ and translation $\bm{t}_b \in \mathbb{R}^2$.
After posing the model and recovering the posed joints $X$, we project them on the image:
\begin{equation}
    \label{eq:weak-persp}
    \bm{x} = s(\Pi(X) + \bm{t})
\end{equation}
\noindent where $\Pi$ is an orthographic projection.
We then compute a bounding box for each hand and the face, from the corresponding subsets of projected \twoD
joints, $\bm{x}_h$ and $\bm{x}_f$.
Let $(x_\text{min}, y_\text{min})$ and $(x_\text{max}, y_\text{max})$ be the top left and
bottom right points for a part, computed from the respective joints.
The bounding box center is equal to
$\bm{c}=\left(\frac{x_\text{min} + x_\text{max}}{2},\frac{y_\text{min} + y_\text{max}}{2}\right)$,
and its size is $b_s=2 \cdot \text{max}(x_\text{max} - x_\text{min}, \allowbreak y_\text{max} - y_\text{min})$.
Using these boxes, we compute affine
transformations $T_{h}$, $T_{f} \in \mathbb{R}^{2 \times 3}$ to extract higher resolution hand and faces images
using spatial transformers (ST)~\cite{jaderberg2015spatial}:
\begin{align}
    & I_h = \text{ST}\left(I; T_h\right), \, I_f = \text{ST}\left(I; T_f\right).\label{eq:part_crops}
\end{align}
The hand $I_h$ and face $I_f$ images are fed to a hand network $h$ and a face network $f$,
to refine the respective parameter predictions. Hand parameters $\pose_h$
include the orientation of the wrist $\pose^{\text{wrist}}$
and finger articulation $\pose^\text{fingers}$,
while face parameters contain the expression coefficients $\expression_f$ and
facial pose $\pose_f$, which is just the rotation of the jaw.
We refine the parameters of the body network by predicting offsets for each of the parameters and condition the part specific networks
on the corresponding body parameters:%
\begin{equation}
     [\Delta\pose^{\text{wrist}}, \Delta\pose^\text{fingers}] = h\left(I_h;\pose_b^{\text{wrist}}, \pose_b^\text{fingers}\right), \,
    \left[\Delta\pose_{\text{f}}, \Delta\expression \right] =
    f\left(I_f; \pose_b^{\text{f}}, \expression_b\right)
\end{equation}
\noindent where $\pose_b^{\text{wrist}}$, $\pose_b^\text{fingers}$,
$\pose_b^{\text{f}}$, $\expression_b$
are the wrist pose, finger pose, facial pose and expression predicted by $g(\cdot)$.
The hand and head sub-networks also produce a set of weak-perspective camera parameters $\left\{s_\text{h}, \bm{t}_\text{h}\right\}$,
$\left\{ s_\text{f}, \bm{t}_\text{f}\right\}$ that align the predicted \threeD meshes to their respective images
$I_h$ and $I_f$. The final hand and face predictions are then equal to:
\begin{align}
    \pose_h =\left[\pose^\text{wrist}, \pose^\text{fingers}\right] &=
    \left[\pose_b^\text{wrist}, \pose_b^\text{fingers}\right] +
    \left[\Delta\pose_{\text{wrist}}, \Delta\pose_\text{fingers}\right]
    \\
     \left[\expression, \pose_f \right] &=
        \left[\expression_b, \pose_b^\face \right]  +
        \left[\Delta\expression, \Delta\pose_f \right].
\end{align}%
With this approach we can utilize the full resolution of the original image $I$
to overcome the small pixel resolution of the hands and face in the body image $I_b$.
Another significant advantage is that we are able to leverage hand- and face-only data
to supplement the training of the hand and face sub-networks.
A detailed visualization of the prediction process can be seen in Figure~\ref{fig:arch}.
The loss function used to train the model is a combination of terms for the
body, the hands and the face. We train the body network using a combination of a \twoD re-projection loss, \threeD joint errors and a loss on the parameters $\Theta$,
when available. All variables with a hat denote \groundtruth quantities.
\begin{align}
     L 												&= L_{\text{body}} + L_{\text{hand}} + L_{\text{face}} + L_h + L_f\label{eq:total_loss} \\
     L_{\text{body}} 								&= L_{\text{reproj}} + L_{\text{\threeD Joints}} + L_{\text{\smplx}} \\
     L_{\text{\threeD Joints}} + L_{\text{\smplx}} 	&= \sum_{j=1}^J \normabs{\hat{\bm{X}}_j - \bm{X}_j} + \normmse{\left\{\hat{\shape}, \hat{\pose},\hat{\expression}\right\} - \left\{\shape, \pose, \expression\right\} } \label{eq:body_loss} \\
     L_{\text{reproj}} 								&= \sum_{j=1}^J\upsilon_j \normabs{\hat{\bm{x}}_j - \bm{x}_j}. \label{eq:reproj}
\end{align}
We use $\upsilon_j$ as a binary variable denoting visibility of each of the $J$ joints.
The re-projection losses $L_h$ and $L_f$
are applied in the hand and face image coordinate space,
using the affine transformations $T_h$, $T_f$. The reason for this extra penalty is that
alignment errors in the \twoD projection of the fingers or the facial landmarks have a much smaller magnitude compared
to those of the main body joints when computed on the body image $I_b$
\begin{align}
    & L_{h} = \sum_{j \in \text{Hand}}\upsilon_j \normabs{T_h T_b^{-1} \left(\hat{\bm{x}}_j - \bm{x}_j\right)}, \,
    L_{f} = \sum_{j \in \text{Face}}\upsilon_j \normabs{T_f T_b^{-1} \left(\hat{\bm{x}}_j - \bm{x}_j\right)}.
\end{align}

For the hand and head only data we also employ a re-projection loss, using only
the subset of joints of each part, and parameter losses:
\begin{align}
    L_{\text{hand}} &= L_{\text{reproj}} + \normmse{
        \left\{\hat{\shape}_h, \hat{\pose}_h\right\} - \left\{\shape_h, \pose_h\right\} } \label{eq:hand_loss} \\
    L_{\text{face}} &= L_{\text{reproj}} + \normmse{
    \left\{\hat{\shape}_f, \hat{\pose}_f, \hat{\expression}_f\right\} - \left\{\shape_f, \pose_f, \expression_f\right\} }.\label{eq:face_loss}
\end{align}

\subsection{Implementation Details}
\label{subsection:impl}

\newcommand{\trainDataSizL}{0.80}
\newcommand{\trainDataSizR}{0.40}
\begin{figure}[t]
    \centering
    \begin{subfigure}[!h]{0.32\textwidth}
        \centering
        \includegraphics[trim=070mm 000mm 060mm 020mm, clip=true, height=\trainDataSizL \linewidth]{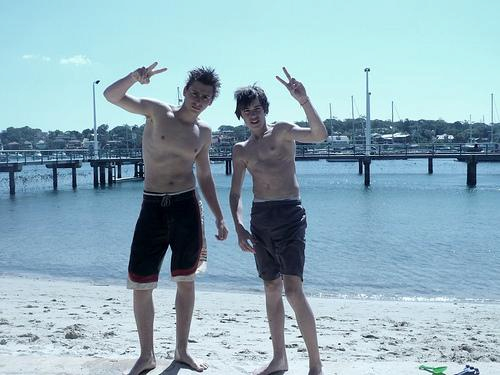}%
        \includegraphics[trim=070mm 000mm 060mm 020mm, clip=true, height=\trainDataSizL \linewidth]{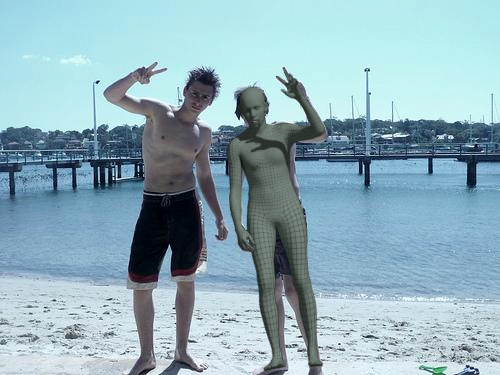}
        \phantomsubcaption
    \end{subfigure}
    \begin{subfigure}{0.32\textwidth}
        \centering
        \includegraphics[trim=000mm 000mm 000mm 000mm, clip=false, height=\trainDataSizR \linewidth]{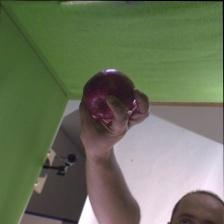}%
        \includegraphics[trim=000mm 000mm 000mm 000mm, clip=false, height=\trainDataSizR \linewidth]{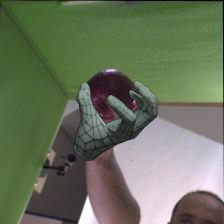}\\
        \includegraphics[trim=000mm 000mm 000mm 000mm, clip=false, height=\trainDataSizR \linewidth]{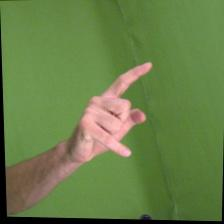}%
        \includegraphics[trim=000mm 000mm 000mm 000mm, clip=false, height=\trainDataSizR \linewidth]{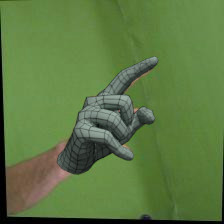}
        \phantomsubcaption
    \end{subfigure}
    \begin{subfigure}{0.32\textwidth}
        \centering
        \includegraphics[trim=000mm 000mm 000mm 000mm, clip=false, height=\trainDataSizR \linewidth]{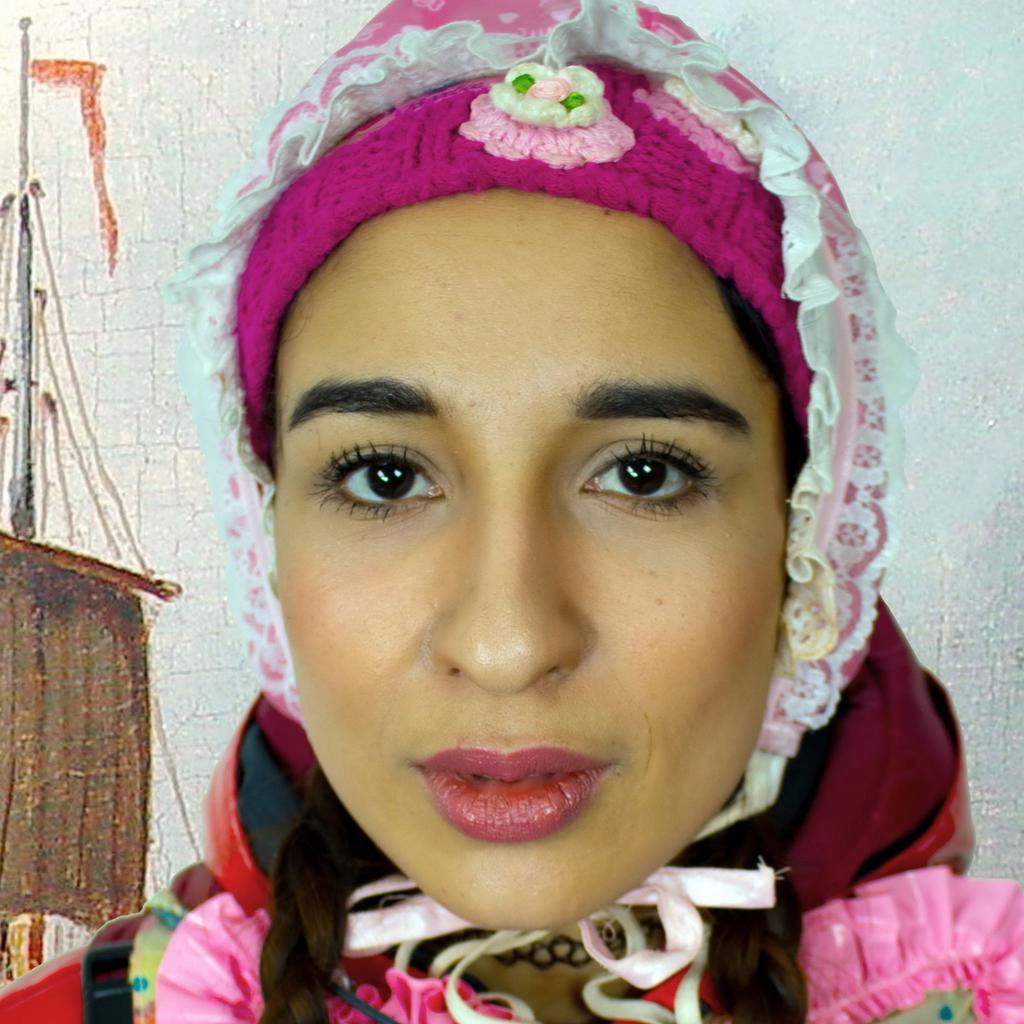}%
        \includegraphics[trim=000mm 000mm 000mm 000mm, clip=false, height=\trainDataSizR \linewidth]{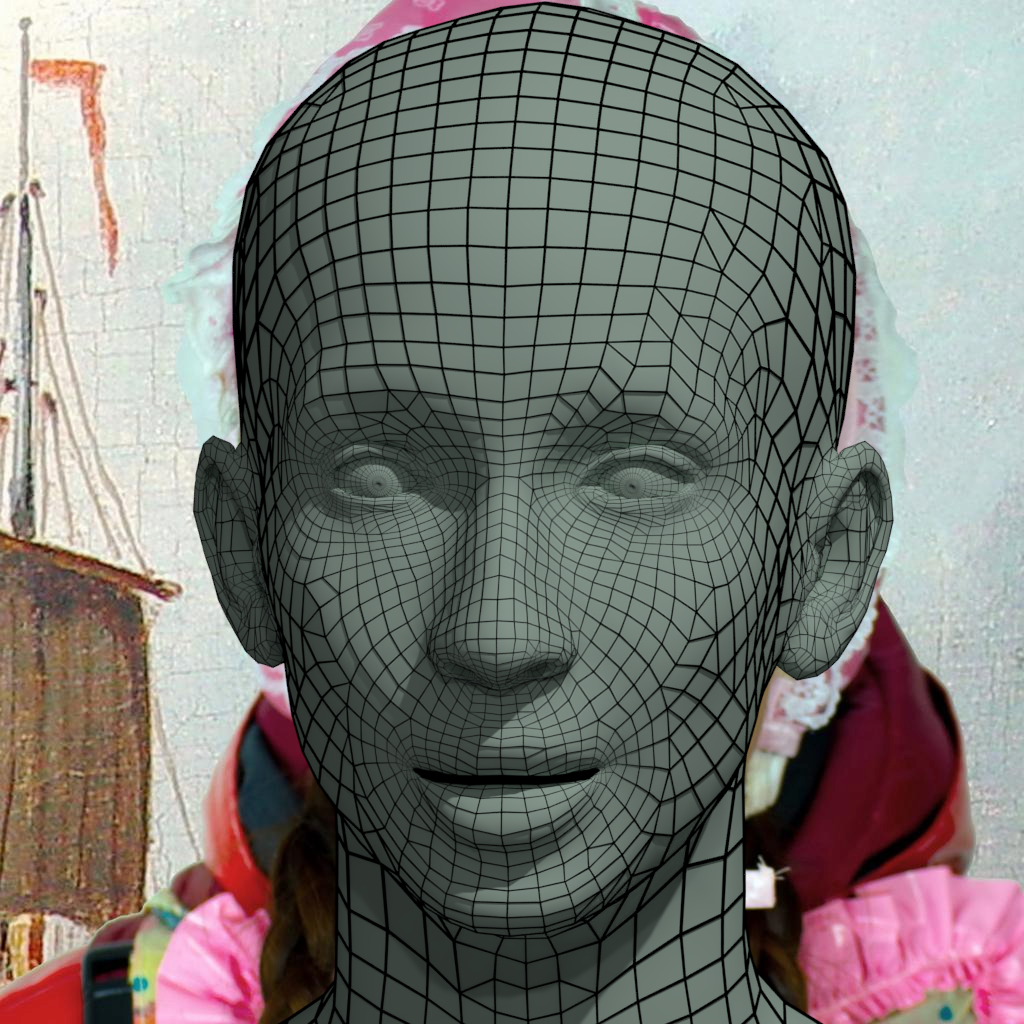}\\
        \includegraphics[trim=000mm 000mm 000mm 000mm, clip=false, height=\trainDataSizR \linewidth]{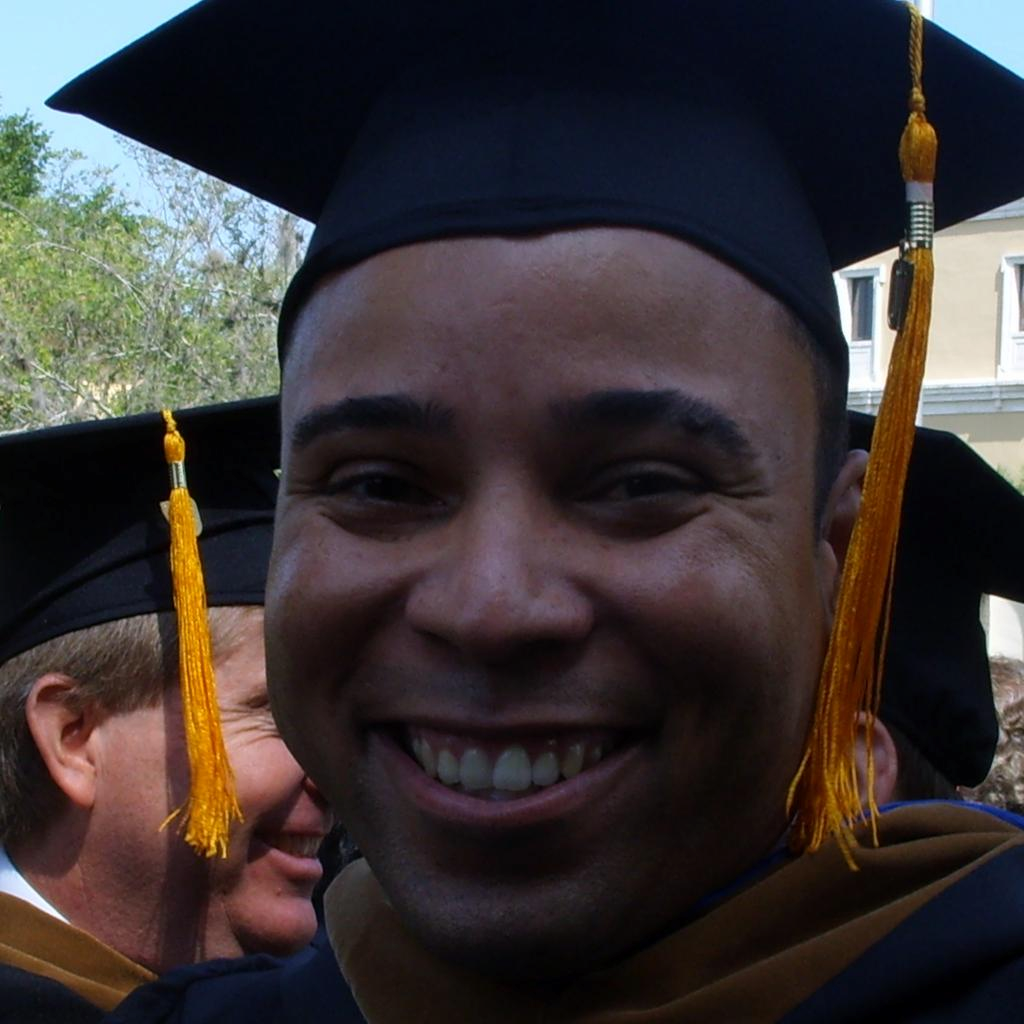}%
        \includegraphics[trim=000mm 000mm 000mm 000mm, clip=false, height=\trainDataSizR \linewidth]{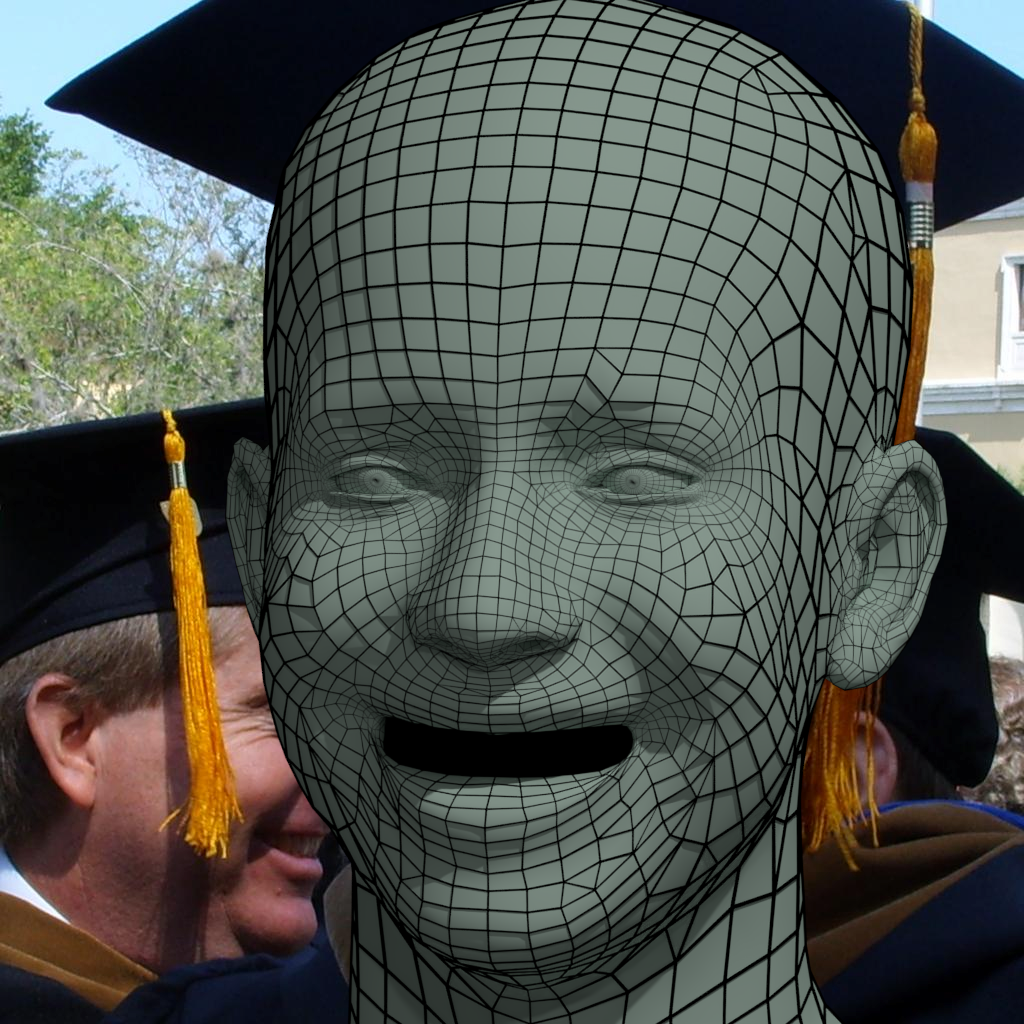}
        \phantomsubcaption
    \end{subfigure}
    \caption{
        \labelLEFT Example  curated expressive fit.
        \labelMIDDLE Hands sampled from the \freihand dataset~\cite{Freihand2019}.
        \labelRIGHT Head training data produced by running \ringnet~\cite{sanyal_2019_cvpr}
        on FFHQ~\cite{karras2019style} and then fitting to \twoD landmarks predicted by~\cite{bulat2017far}.
    }
    \label{fig:training_samples}%
\end{figure}

\textbf{Datasets}:
We curate a dataset of \smplx fits by running vanilla \smplifyx \cite{Pavlakos_2019_CVPR} on the LSP~\cite{johnson2010clustered}, LSP extended~\cite{johnson2011learning} and MPII~\cite{andriluka20142d} datasets.
We then ask human annotators whether the resulting body mesh is plausible and agrees with the image and collect $32,617$ pairs of images and \smplx parameters.		%
To augment the training data for the body we transfer the public fits of SPIN
\cite{Kolotouros_2019_ICCV} from \smpl to \smplx, see \supmat
Moreover, we use H3.6M~\cite{ionescupapavaetal2014} for additional \threeD supervision for the body.
For the hand sub-network we employ the hand-only data of \freihand~\cite{Freihand2019}.
For the face sub-network we create a \pseudogt face dataset by running \ringnet~\cite{sanyal_2019_cvpr} on FFHQ~\cite{karras2019style}.
The regressed \flame~\cite{FLAME:SiggraphAsia2017} parameters are refined by fitting to facial landmarks~\cite{bulat2017far} for better alignment with the image and more detailed expressions.
Figure~\ref{fig:training_samples} shows samples from all training datasets.

\textbf{Architecture}:
For the body network we extract features $\phi \in \mathbb{R}^{2048}$ with HRNet~\cite{SunXLW19}.
For the face and hand sub-networks we use a ResNet18 \cite{he2016deep} to limit the computational cost.
For all networks, rather than directly regressing the parameters
$\Theta$ from $\phi$, we follow the iterative process of~\cite{kanazawa_cvpr_2018}.
We start from an initial estimate $\Theta_0 = \bar{\Theta}$, where $\bar{\Theta}$ represents the mean,
which is concatenated to the features $\phi$ and fed to an MLP that predicts a residual
$\Delta\Theta_1 = \text{MLP}\left(\left[\phi, \Theta_0\right]\right)$. The
new parameter value is now equal to $\Theta_1 = \Theta_0 + \Delta\Theta_1$ and
the whole process is repeated. As in~\cite{kanazawa_cvpr_2018}, we iterate for $t=3$ times.
The entire pipeline is implemented in \pytorch~\cite{pytorch}.
For architecture details see \supmat

\textbf{Data Pre-processing and Augmentation}:
We follow the pre-processing and augmentation protocol of \cite{Kolotouros_2019_ICCV} for all networks.
To make the model robust to partially visible bodies we adopt the cropping augmentation of Joo \etal~\cite{joo2020exemplar}.
In addition, we augment the hand- and face-only images with random translations, as well as down-sampling by a random factor and then up-sampling back to the original resolution.
The former simulates %
a misaligned body prediction, while the latter bridges the gap in image quality between the full-body and part-specific data.
Hand and especially face images usually have a much higher resolution and quality compared to a crop extracted from a full-body image.
To simulate body conditioning for the hand- and head-only data we add random noise to the initial point of the iterative regressor.
For the hands we replace the default finger pose with a random rotation $r_{\text{finger}}$ sampled from the PCA pose space of \mano. %
For the head  we replace the default jaw rotation $\bar{\pose}_f$ with a random rotation of $r_{\text{f}} \sim \mathcal{U}\left(0, 45\right)$ degrees around the x-axis. %
For both parts, we replace their global rotation with a random rotation with angle $r_{\text{global}} \sim \mathcal{U}\left(r_{\text{min}}, r_{\text{max}}\right)$
and the same axis of rotation as the corresponding \groundtruth.
We set
$\left(r_{\text{min}}, r_{\text{max}}\right)_{\text{hand}}$ to $(-90, 90)$ and
$\left(r_{\text{min}}, r_{\text{max}}\right)_{\text{face}}$ to $(-45, 45)$ degrees.
The default mean shape is replaced with a random vector
$\shape \sim \mathcal{N}\left(\bm{0}, I\right), I \in \mathbb{R}^{10\times 10}$ and the default neutral expression with a random expression $\expression \sim \mathcal{N}(0, \mathcal{I})$.
Some visualizations of the different types of data augmentation can be found in \supmat

\textbf{Training:}
We first pre-train the body, hand and face networks separately, using ADAM
\cite{adam}, on the respective part-only datasets.
We then fine-tune all networks jointly on the union of all training data, following Section~\ref{subsec:model_training},
letting the network make even better use of the conditioning (see Sec. \ref{sec:experiments} and Tab.~\ref{table:ehf_ablative}).
Please note that for this fine-tuning, our new dataset of curated \smplx fits plays an instrumental role.
Our exact hyper-parameters are included in the released training code.

\section{Experiments}
\label{sec:experiments}

\subsection{Evaluation Datasets}
\label{subsec:eval_datasets}
We evaluate on several datasets:

\textbf{Expressive Hands and Faces (\ehf)}~\cite{Pavlakos_2019_CVPR}
consists of $100$ \rgb images paired with \smplx registrations to synchronized \threeD scans.
It contains a single subject performing a variety of interesting body poses, hand gestures and facial expressions.
We use it to evaluate our whole-body predictions.

\textbf{\threeD Poses in the Wild (3DPW)}~\cite{vonmarcard_eccv_2018_3dpw}
consists of \inthewild \rgb video sequences annotated with \threeD \smpl poses.
It contains several actors performing various motions, in both indoor and outdoor environments.
It is captured using a single \rgb camera and IMUs mounted on the subjects.
We use it to evaluate our predictions for the main body area, excluding the head and hands.

\textbf{\freihand}~\cite{Freihand2019}
is a multi-view \rgb hand dataset that contains \threeD \mano hand pose and shape annotations.
The \groundtruth for the test data is held-out and evaluation is performed by submitting the estimated hand meshes to an online server.
We use it to evaluate our hand sub-network predictions. %

\textbf{Stirling/ESRC 3D}~\cite{feng2018evaluation}
consists of facial \rgb images with \groundtruth \threeD face scans.
It contains $2000$ neutral faces images, namely $656$ high-quality (HQ) ones and $1344$ low-quality (LQ) ones.
We use it to evaluate our face sub-network following the protocol of~\cite{feng2018evaluation}.

\subsection{Evaluation Metrics}
\label{subsec:metrics}
We employ several common metrics below.
We report errors with and without rigid alignment to the \groundtruth.
A ``PA'' prefix denotes that the metric measures error after solving
for  rotation, scale and translation using Procrustes Alignment.

To compare with \groundtruth \threeD skeletons, we use the \textbf{Mean Per-Joint Position Error (MPJPE)}.
For this, we first compute the $14$ LSP-common joints, by applying a linear joint regressor on the \groundtruth
and estimated meshes, and then compute their mean Euclidean distance.

For comparing to \groundtruth meshes, we use the \textbf{Vertex-to-Vertex (V2V)}
error, \ie the mean distance between the \groundtruth and predicted mesh vertices.
This is appropriate when the predicted and \groundtruth meshes have the same topology, \eg
\smplx 	for our overall network,
\mano 	for our hand and
\flame 	for our face sub-network.
For a fair comparison to methods that predict \smpl instead of \smplx,
like \cite{kanazawa_cvpr_2018,Kolotouros_2019_ICCV}, we also report V2V only on the main-body,
\ie without the hands and the head,
as \smpl and \smplx share common topology for this subset of vertices.

For comparing to approaches that output meshes with different topology, like MTC~\cite{Xiang_2019_CVPR} that uses the \adam model and not \smplx, we cannot use V2V. %
Instead, we compute the (mesh-to-mesh) \textbf{point-to-surface} (P2S)
distance from the \groundtruth mesh, as a common reference, to the estimated mesh.

For evaluation on datasets that include \groundtruth scans, we compute a \textbf{scan-to-mesh} version of the above \textbf{point-to-surface} distance,
namely from the \groundtruth scan points to the estimated mesh surface.
We use this for the face dataset of~\cite{feng2018evaluation} to evaluate the head estimation of our face sub-network.

Finally, for the \freihand dataset \cite{Freihand2019} we report all metrics returned by their evaluation server.
Apart from PA-MPJPE and PA-V2V described above, we also report the \textbf{F-score}~\cite{knapitsch2017tanks}.

\subsection{Quantitative and Qualitative Experiments}
\label{subsec:main_experiments}

First, we evaluate our approach on the 3DPW dataset that includes \smpl \groundtruth meshes.
Although this does not include \groundtruth hands and faces, it is ideal for comparing main-body reconstruction against state-of-the-art approaches, namely HMR~\cite{kanazawa_cvpr_2018} and SPIN~\cite{Kolotouros_2019_ICCV}.
Table~\ref{table:pose_results_3dpw} presents the results, and shows that \modelname outperforms HMR and is on par with the more recent SPIN.
This confirms that \modelname provides a solid foundation upon which to build
detailed reconstruction for the hands and face.

\begin{table}[t!]
    \centering
    \caption{
        Comparison on the 3DPW dataset~\cite{vonmarcard_eccv_2018_3dpw}
        with two state-of-the-art approaches for \smpl regression,
        HMR~\cite{kanazawa_cvpr_2018} and SPIN~\cite{Kolotouros_2019_ICCV}.
        The numbers are per-joint and per-vertex errors (in mm) for the body
        part of \smpl.
        \modelname outperforms HMR and is on par with SPIN, while also
        being able to capture details for the hands and the face.
    }
    \scriptsize
    \begin{tabular}{l|c|c|c}
        \toprule
        Method & PA-MPJPE (mm) & MPJPE (mm) & PA-Body V2V (mm) \\
        \toprule
        HMR    \cite{kanazawa_cvpr_2018}  & 81.3 &  130  &  65.2 \\
        SPIN    \cite{Kolotouros_2019_ICCV} & 59.2  & 96.9 & 53.0 \\
        \modelname & 60.7  & 93.4 &  55.6 \\
        \bottomrule
    \end{tabular}
    \label{table:pose_results_3dpw}
\end{table}

We then evaluate on the \ehf dataset that includes high-quality \smplx ground truth.
This allows evaluation for the more challenging task of holistic body reconstruction, including expressive hands and face.
Table~\ref{table:ehf_ablative} presents an ablation study for our main components.
In the first row, we see that the initial body network, that uses a low-resolution body-crop image as input,
performs well for body reconstruction but makes mistakes with the
hands. %
\highlightCR{The next two rows add {\it body-driven attention}; they use the body network prediction to locate the hands and face,
and then redirect the attention in the original image, crop higher-resolution image patches for them, and feed them to the respective hand and face sub-networks to refine their predictions,
while initializing/conditioning their predictions.}
\highlightCR{This conditioning can take place in two ways.}
\highlightCR{The second row shows a naive combination using independently trained sub-networks.}
This fails to significantly improve the results,
since there is a domain gap between images of face- or hand-only~\cite{feng2018evaluation,Freihand2019} training datasets and
hand/head image crops from full-body \cite{andriluka20142d,johnson2010clustered,johnson2011learning} training datasets;
\highlightCR{the former tend to be of higher resolution and better image
quality.} %
\highlightCR{Please note that this is similar to \cite{cao2018openpose}, but extended for \threeD mesh regression.}
\highlightCR{In the third row, the entire pipeline is fine-tuned
  end-to-end.}
\highlightCR{This results in a boost in quantitative performance,
  improving mainly hand articulation (best overall performance).} %

\begin{table}[t!]
    \centering
    \caption{Ablation study on the EHF dataset.
        The results are vertex-to-vertex errors expressed in mm for the different parts
        (i.e., all vertices, body vertices, hand vertices and head vertices).
        We report results for the initial body network applied on the low resolution (first row),
        for a version that uses the body-driven attention to estimate hands and faces (second row),
        and for the final regressor that jointly fine-tunes the body, hands and face sub-networks.
    }
    \scriptsize
    \begin{tabular}{l|c|c|c|c|c|c}
        \toprule
        \multirow{2}{*}{Networks} 	&	\multirow{1}{*}{~Attention on~}		& 	~End-to-end~		&  \multicolumn{4}{c}{PA-V2V (mm)} 						\\	\cline{4-7}
        	{}							& 	\multirow{1}{*}{~high-res. crops~}	& 	~fine-tuning~ 	&  All		&	Body		&	L/R hand			&	Face		\\
        \midrule
        Body only	 				& 	\xmark								& 	\xmark			& ~57.3~		&	~55.9~	&	~14.3 / 14.8~	&	~5.8~ 	\\
        Body \& Hand \& Face			& 	\cmark								& 	\xmark			&  56.4		&	 52.6	&	 14.1 / 13.9		&	 6.0		\\
        Body \& Hand \& Face			& 	\cmark								& 	\cmark			&  54.5		&	 52.6	&	 13.1 / 12.5		&	 5.8		\\
        \bottomrule%
    \end{tabular}
    \label{table:ehf_ablative}
\end{table}

Next,   we compare to state-of-the-art approaches again on the EHF dataset.
First, we compare against the most relevant baseline,
\smplifyx~\cite{Pavlakos_2019_CVPR}, which estimates \smplx using an
optimization approach.
Second, we compare against Monocular Total Capture
(\mtc)~\cite{Xiang_2019_CVPR}, which estimates expressive \threeD
humans using the \adam model.
Note that we use their publicly available implementation, which
does not include an expressive face model.
Third,  we compare against HMR~\cite{kanazawa_cvpr_2018} and
\spin~\cite{Kolotouros_2019_ICCV}, which estimate \smpl bodies,
therefore we perform body-only evaluation, excluding the hand and head regions.
We summarize all evaluations in Table~\ref{table:ehf}. %
We find that \modelname outperforms the other baselines, both in terms of full expressive human reconstruction and body-only reconstruction.
\smplifyx performs a bit better locally, \ie for the hands and face, but the full body pose can be inaccurate, mainly due to errors in OpenPose detections.  %
In contrast, our regression-based approach is a bit less accurate locally for the hands and face, but overall it is more robust than \smplifyx. %
The two approaches could be combined, with \modelname replacing the
heuristic initialization of \smplifyx with its more robust estimation;
we speculate that this would improve both the accuracy and the convergence speed of \smplifyx.
Furthermore, \modelname outperforms \mtc across all metrics.
Finally, it is approximately two orders of magnitude faster than both \smplifyx and \mtc, which are both optimization-based approaches.

\begin{table}[t!]
    \centering
    \caption{Comparison with the state-of-the-art approaches on the EHF dataset.
        The metrics are defined in Sec.~\ref{subsec:metrics}.
        For \smplifyx, the results reported in~\cite{Pavlakos_2019_CVPR} (first row)
        are generated using ground truth camera parameters,
        so they are not directly comparable with the other approaches.
        \mtc running time includes calculation of part orientation fields and \adam fitting.
        The regression based
        methods
        require extra processing to obtain input human bounding box. %
        For example, if one uses Mask-RCNN~\cite{He_2017_ICCV} with a
        \resnet-\fpn~\cite{Lin_2017_CVPR} from \detectron~\cite{wu2019detectron2}
        the complete running time of these methods increases by 43 milliseconds.
        All timings were done with a Intel Xeon W-2123 3.60GHz CPU and
        a Quadro P5000 GPU and are for estimating one person.
    }
    \scriptsize
    \begin{tabular}{p{20ex}|c|c|c|c|c|c|c|c|c|}
        \toprule
        \multirow{2}{*}{Method} & \multirow{2}{*}{Time (s)} &
        \multicolumn{4}{c|}{PA-V2V (mm)} &
        \multicolumn{2}{c|}{PA MPJPE (mm)} &
        \multicolumn{2}{c|}{PA P2S (mm)}
        \\
         \cline{3-10}
        & & All & Body & L/R hand & Face & Body Joints & L/R hand & Mean & Median\\
        \midrule
        $\text{\smplifyx}^{\prime}$~\cite{Pavlakos_2019_CVPR}
        & ~40-60 & \textbf{52.9} & 56.37 & \textbf{11.4}/12.6 & \textbf{5.3} &
        73.5 & \textbf{11.9}/13.2 & \textbf{28.9} & 18.1 \\
        \midrule
        HMR \cite{kanazawa_cvpr_2018} & 0.06 & \na  & 67.2 & \na & \na & 82.0 &
        \na & 34.5 & 21.5 \\
        SPIN \cite{Kolotouros_2019_ICCV} & 0.01  & \na  & 60.6 & \na & \na &
        102.9 & \na & 40.8 & 28.7  \\
        \smplifyx \cite{Pavlakos_2019_CVPR} & ~40-60 & 65.3 & 75.4 & 11.6/12.9 &
        6.3 & 87.6 & 12.2/13.5 & 36.8 & 23.0  \\
        \mtc~\cite{Xiang_2019_CVPR} & 20 & 67.2 & \na & \na & \na &107.8 & 16.3/17.0 & 41.3 & 29.0\\% & 49.3  \\
        \modelname (Ours) & 0.16  & 54.5 & \textbf{52.6} & 13.1/\textbf{12.5} &
        5.8 & \textbf{62.8} & 13.5/\textbf{12.7} & \textbf{28.9} & \textbf{18.0}\\
        \bottomrule%
    \end{tabular}
    \label{table:ehf}
\end{table}

We also evaluate each sub-network on the corresponding part-only datasets.
For the hands we evaluate on the \freihand dataset \cite{Freihand2019}, and for faces on the Stirling/ESRC 3D dataset \cite{feng2018evaluation}.
Table~\ref{table:parts} summarizes all evaluations.
The part sub-networks of \modelname match or come close to the performance of \stateoftheart methods.
We expect that using a deeper backbone, e.g. a \resnet, would be beneficial,
but at a higher computational cost.

\begin{table}[t!]
    \centering
    \caption{We evaluate our final hand sub-network on the \freihand
        dataset~\cite{Freihand2019} and the face sub-network on the test dataset of Feng
        \etal~\cite{feng2018evaluation}. The final part networks are on par with
        existing methods, despite using a shallower backbone,
        \ie a ResNet-18 vs a Resnet-50.%
    }
    \scriptsize
    \begin{tabular}{l|c|c|c|c}
        \toprule
        \freihand & PA-MPJPE (mm) & PA-V2V (mm) & F@5mm & F@15mm \\
        \midrule
        \mano CNN~\cite{Freihand2019} & 11.0 & 10.9 & 0.516 & 0.934  \\
        \modelname hand sub-network $h$ & 12.2 & 11.8 & 0.484 & 0.918 \\
        \midrule%
        Stirling3D Dataset LQ/HQ & Mean (mm) & Median (mm) & \multicolumn{2}{c}{Standard Deviation (mm)}\\
        \midrule
        \ringnet~\cite{sanyal_2019_cvpr} & 2.08/2.02 & 1.63/1.58 & \multicolumn{2}{c}{1.79/1.69}  \\
        \modelname face sub-network $f$ & 2.27/2.42 & 1.76/1.91 & \multicolumn{2}{c}{1.97/2.03} \\
        \bottomrule%
    \end{tabular}
    \label{table:parts}
\end{table}

\begin{figure}[t]
    \centering
    \includegraphics[trim=000mm 014mm 000mm 025mm, clip=true, width=0.28\textwidth]{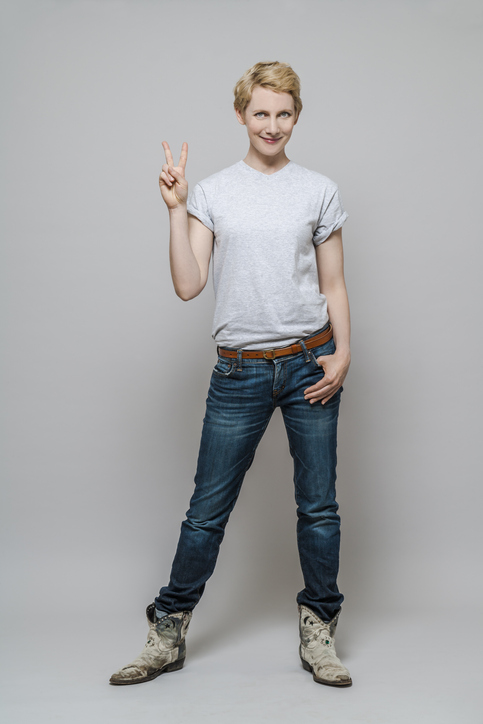}%
    \includegraphics[trim=000mm 014mm 000mm 025mm, clip=true, width=0.28\textwidth]{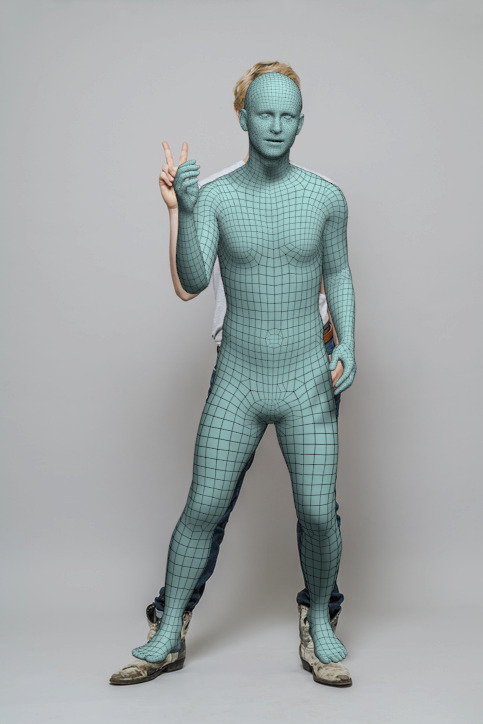}%
    \includegraphics[trim=000mm 014mm 000mm 025mm, clip=true, width=0.28\textwidth]{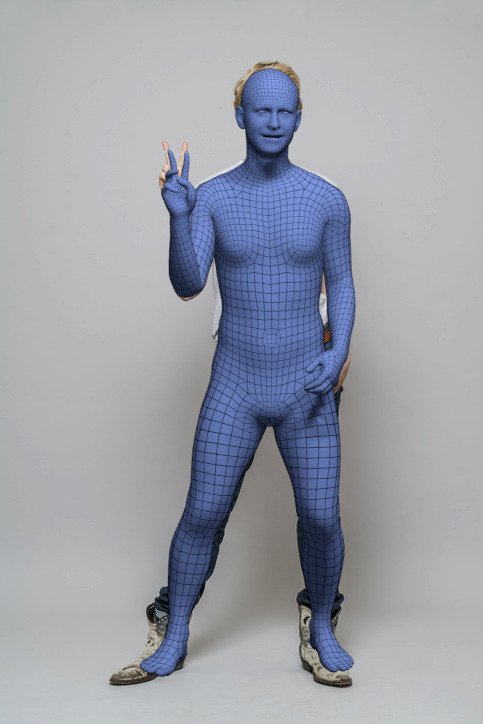}\\%
    \includegraphics[trim=000mm 013mm 000mm 028mm, clip=true, width=0.28\textwidth]{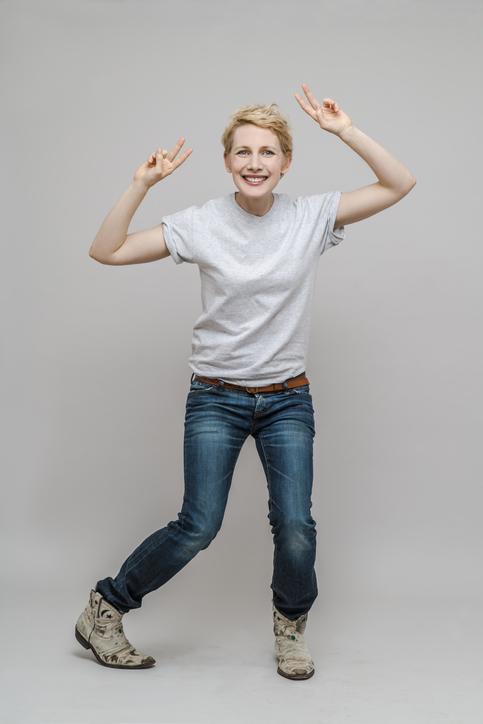}%
    \includegraphics[trim=000mm 013mm 000mm 028mm, clip=true, width=0.28\textwidth]{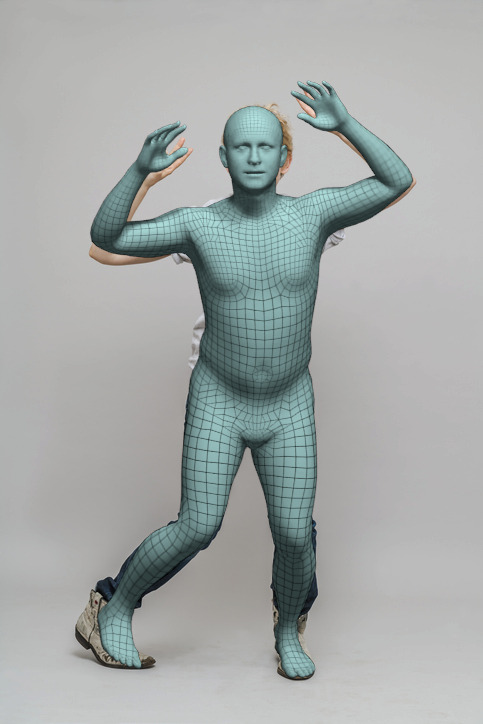}%
    \includegraphics[trim=000mm 013mm 000mm 028mm, clip=true, width=0.28\textwidth]{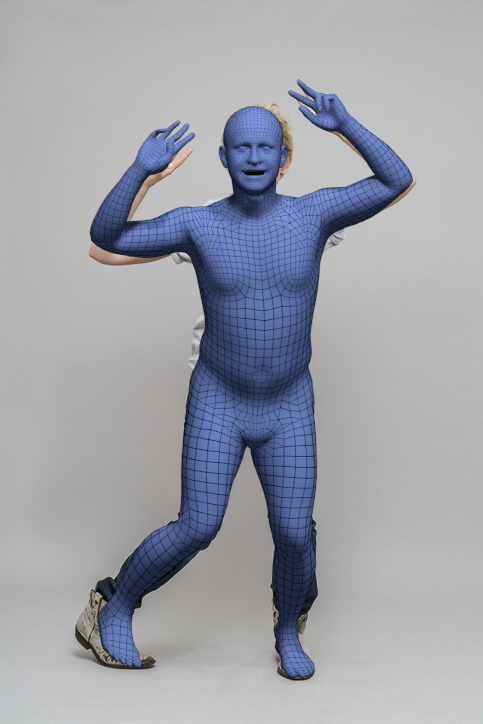}\\

    \caption[]{
        \labelLEFT The input image. %
        \labelMIDDLE Naive regression from a single body image fails to capture
        detailed finger articulation and facial expressions.
        \labelRIGHT \modelname is able
        to recover these details, thanks to its attention mechanism, and
        produces results of similar quality as \smplifyx, while being
        200$\times$ times faster, as seen in Table~\ref{table:ehf}.
    }
    \label{fig:qualitative}%
\end{figure}

\begin{figure}[!t]
    \centering
    \includegraphics[trim=000mm 000mm 000mm 050mm, clip=true, width=0.249\textwidth]{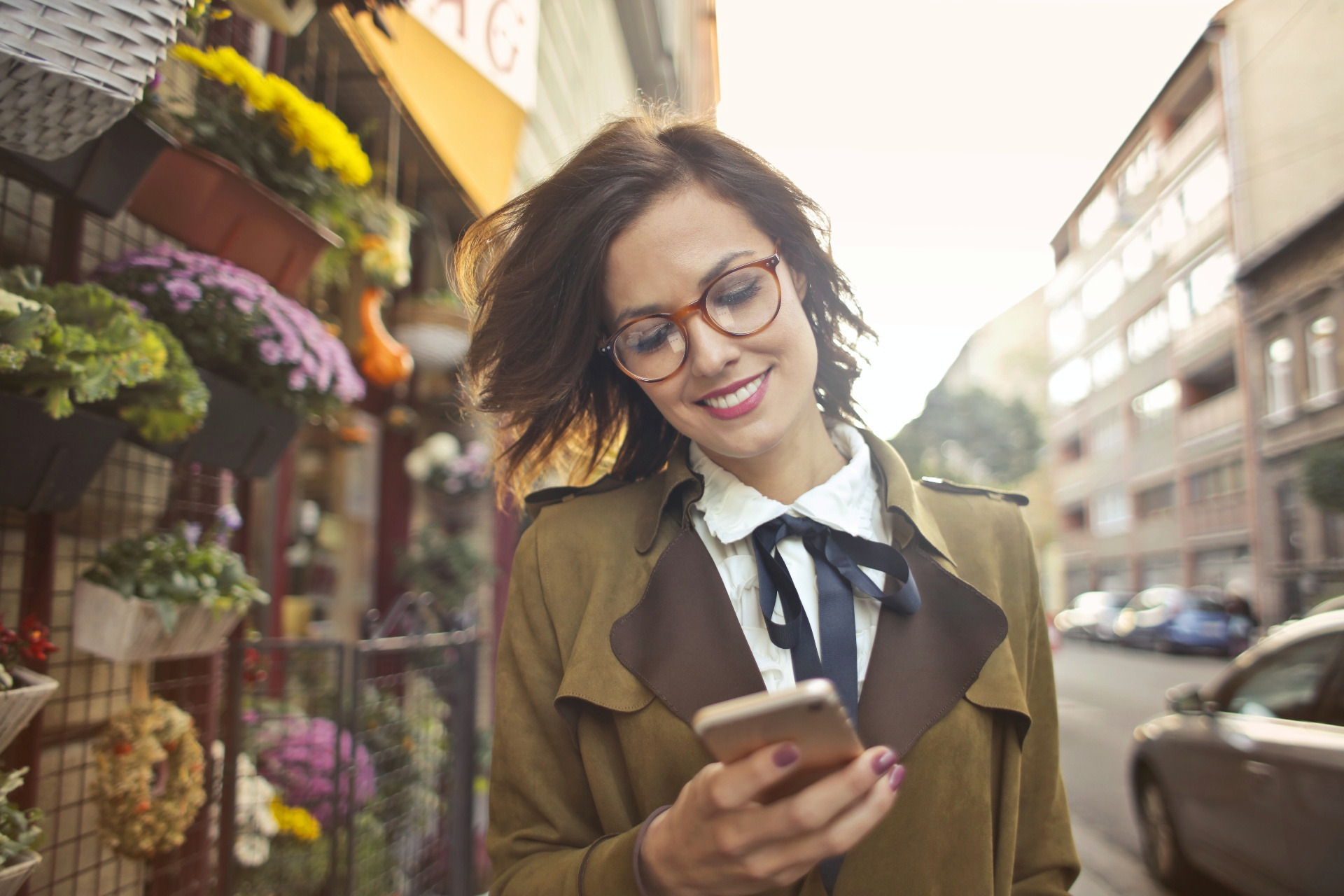}%
    \includegraphics[trim=000mm 000mm 000mm 050mm, clip=true, width=0.249\textwidth]{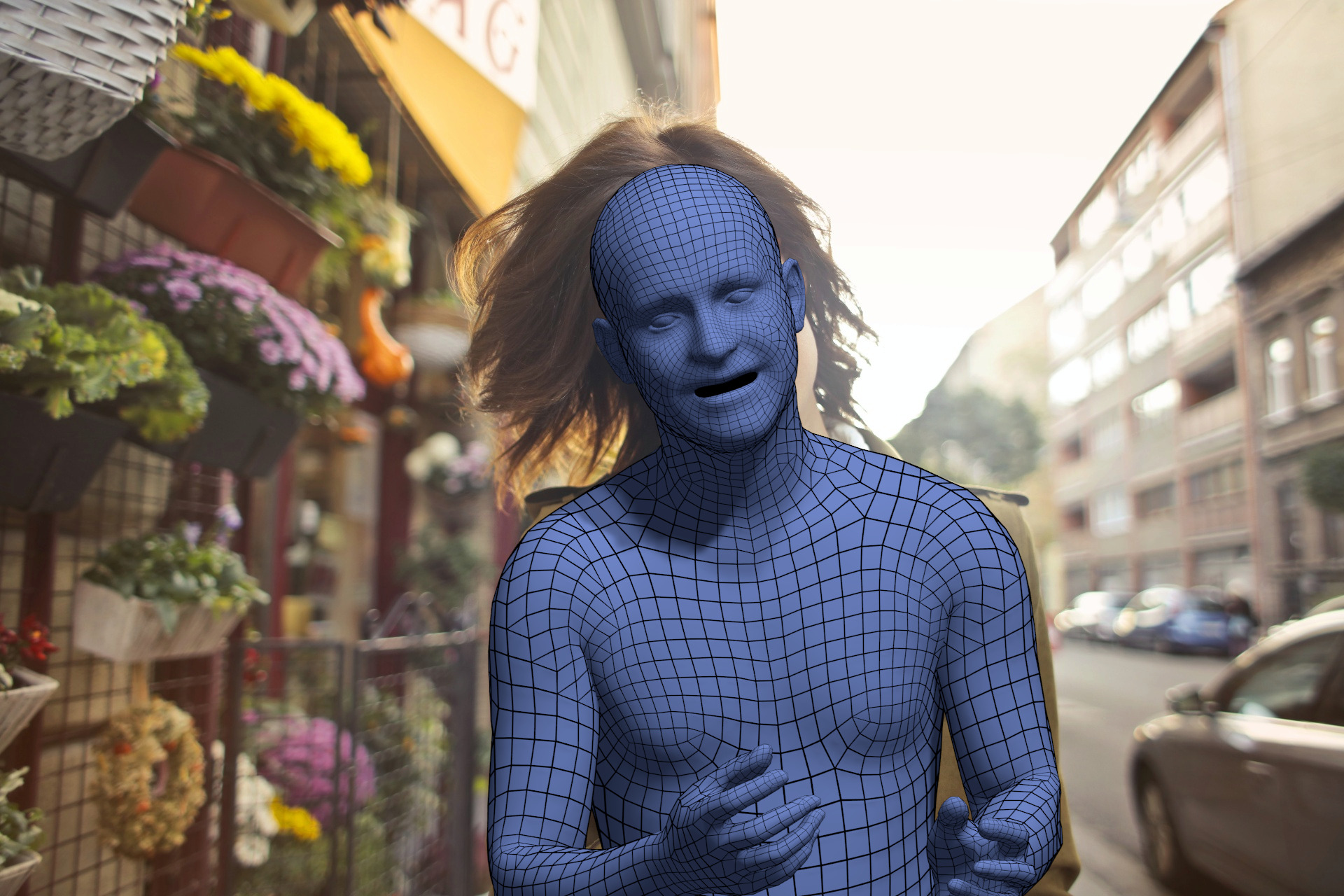}%
    \includegraphics[trim=000mm 000mm 000mm 050mm, clip=true, width=0.249\textwidth]{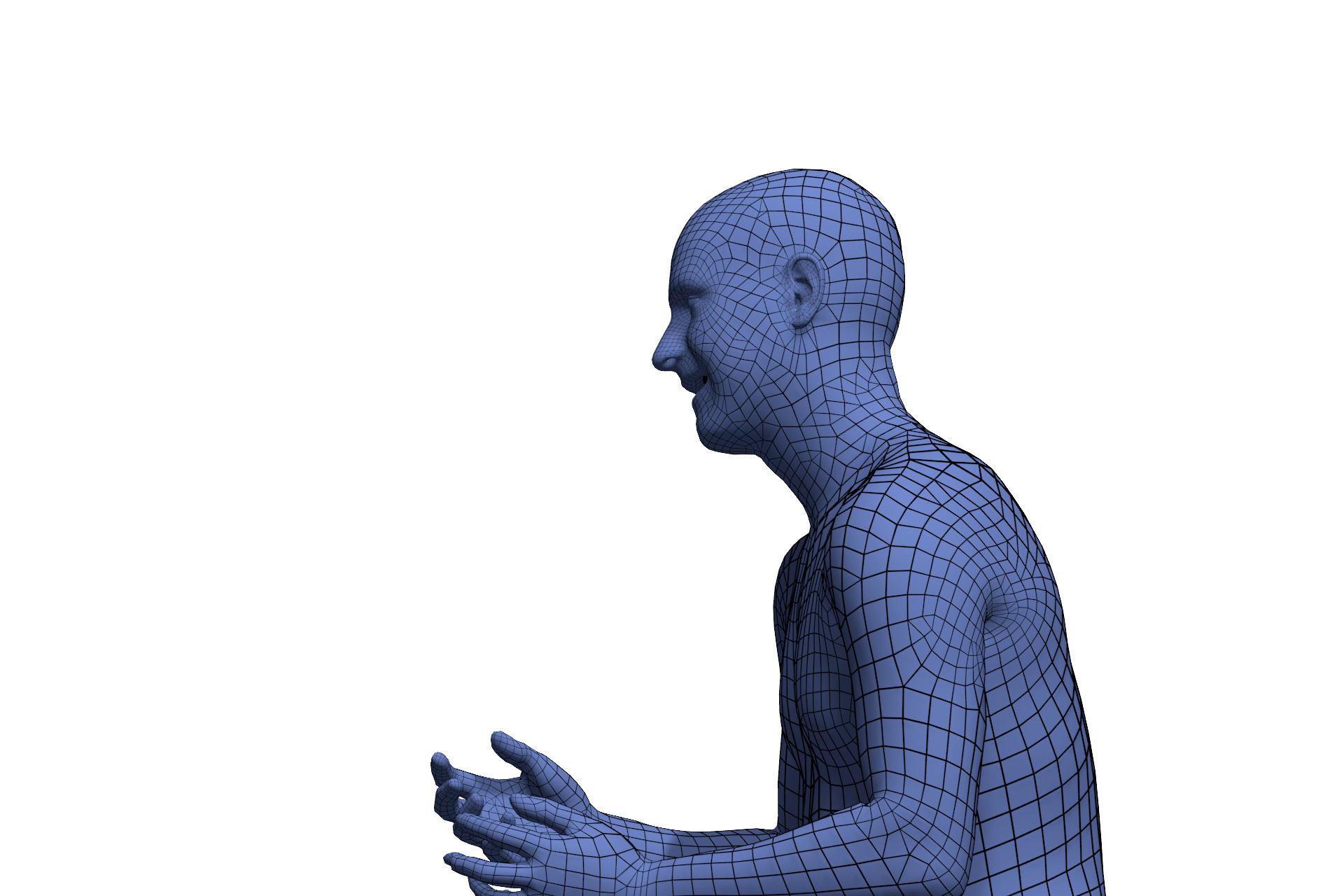}%
    \includegraphics[trim=000mm 000mm 000mm 050mm, clip=true, width=0.249\textwidth]{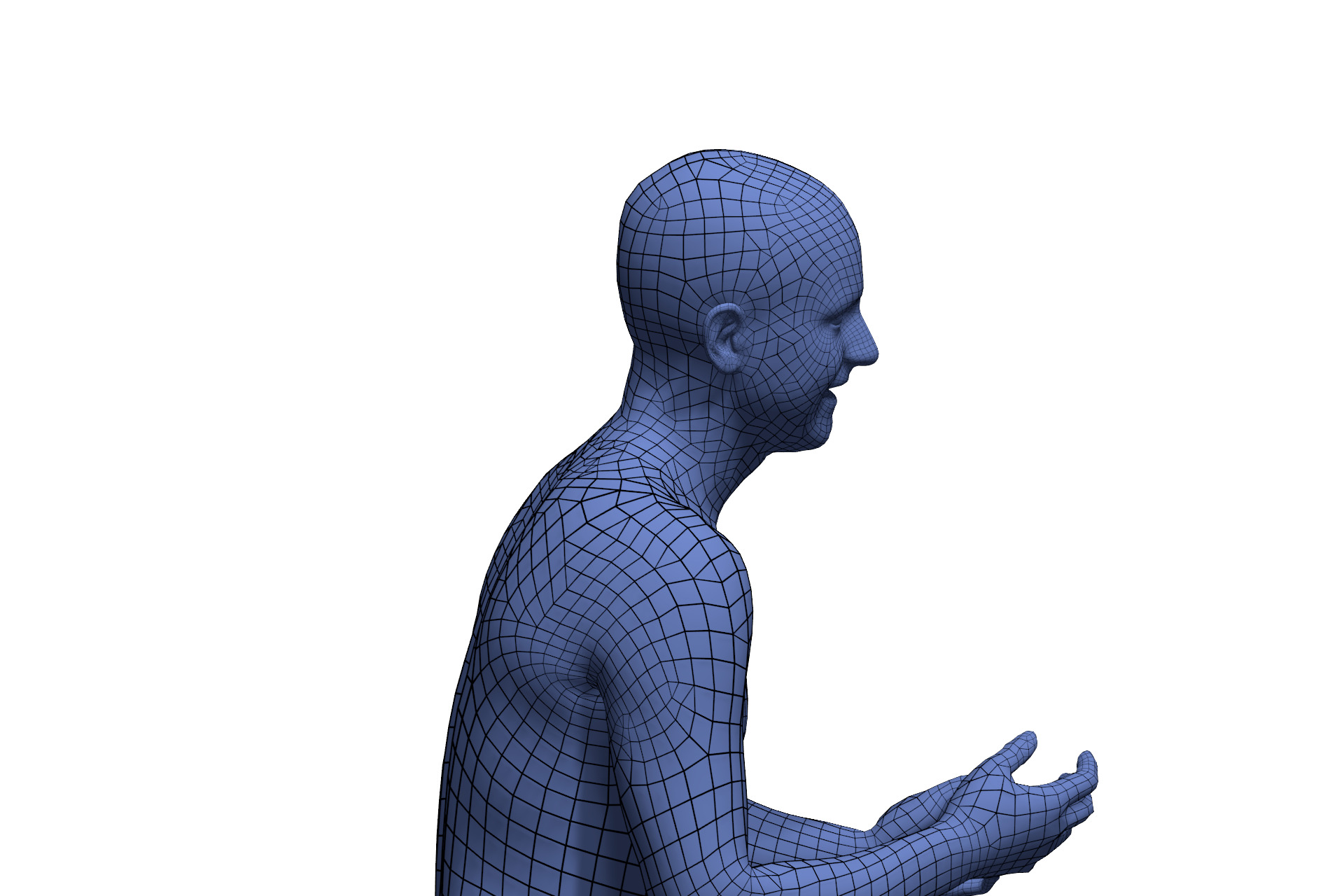}\\%
    \includegraphics[trim=000mm 000mm 000mm 040mm, clip=true, width=0.249\textwidth]{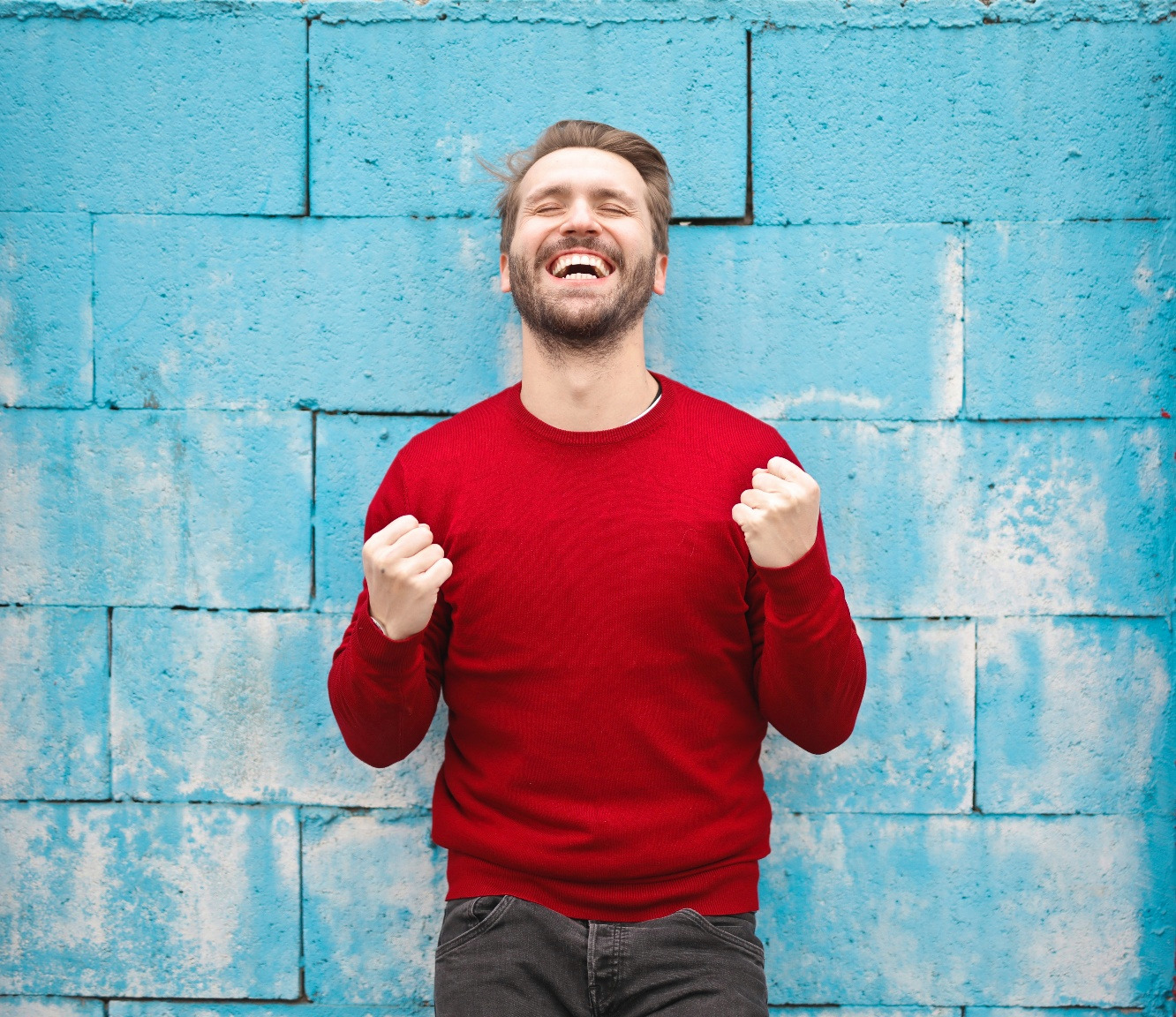}%
    \includegraphics[trim=000mm 000mm 000mm 040mm, clip=true, width=0.249\textwidth]{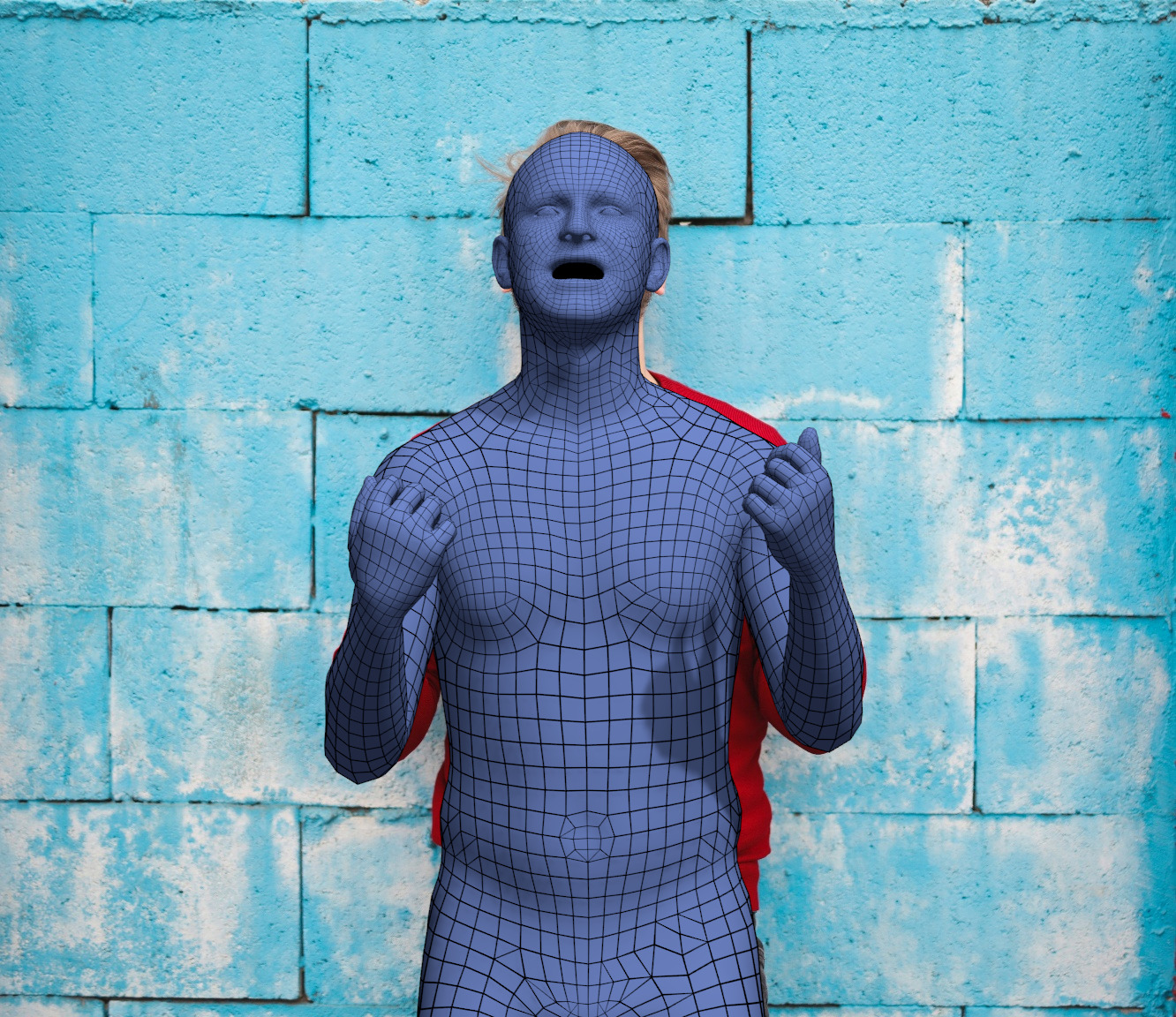}%
    \includegraphics[trim=000mm 000mm 000mm 040mm, clip=true, width=0.249\textwidth]{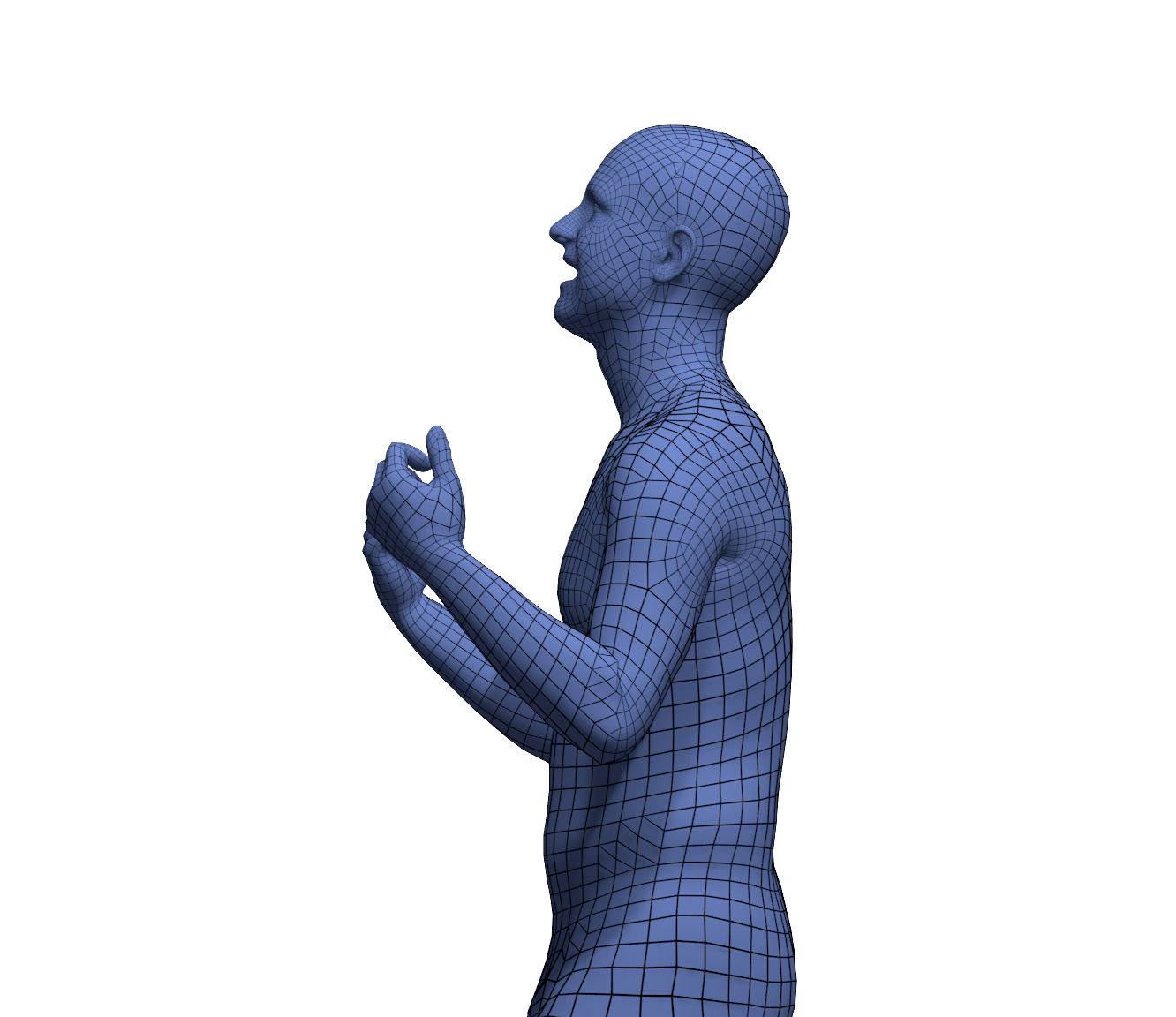}%
    \includegraphics[trim=000mm 000mm 000mm 040mm, clip=true, width=0.249\textwidth]{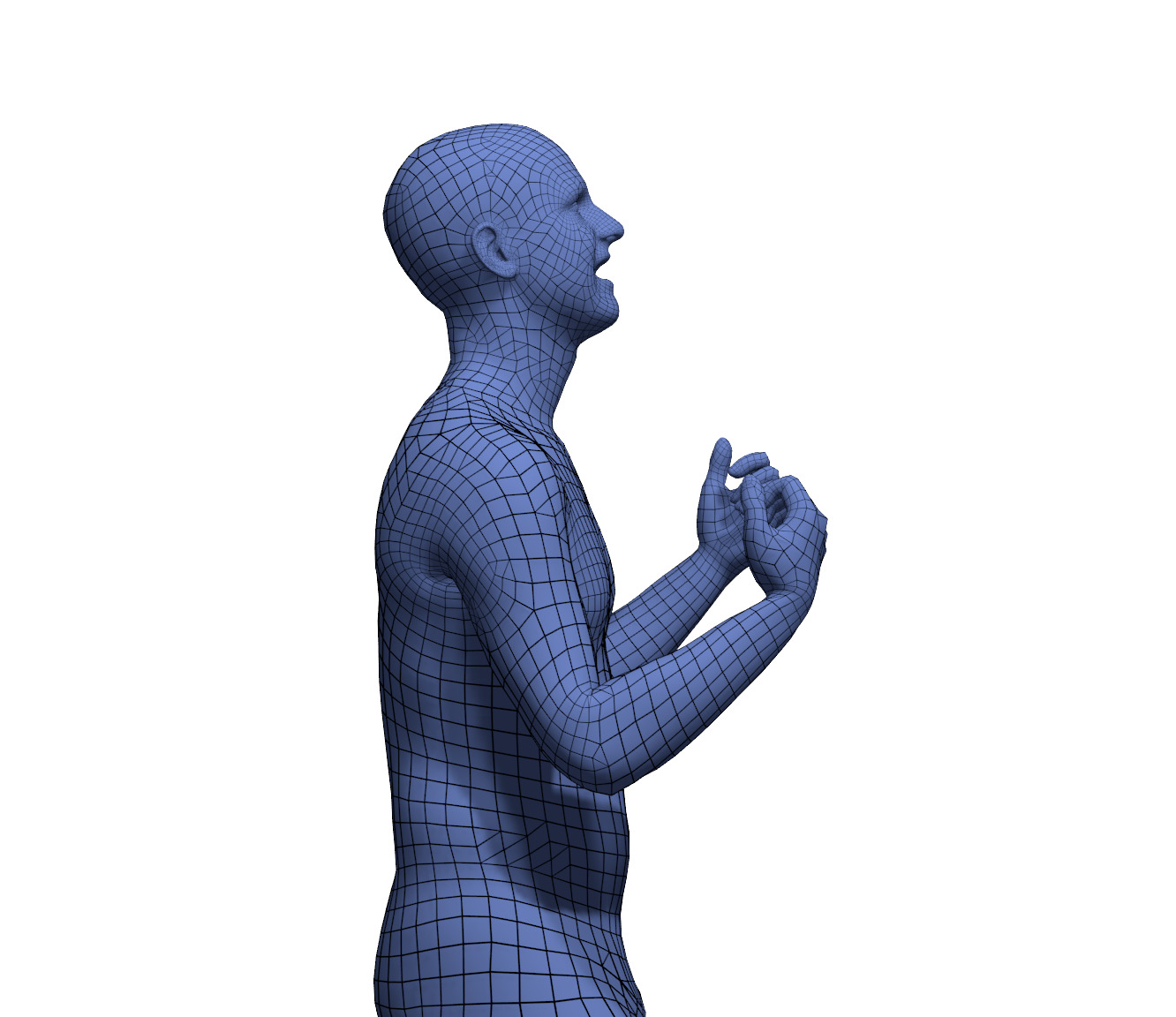}
    \caption[]{
        Input image, \modelname predictions overlayed on the image and
        renderings from different viewpoints.
        \modelname is able to recover detailed hands and faces thanks to its attention mechanism, and
        produces results of similar quality as \smplifyx, while being
        200$\times$ times faster.
    }
    \label{fig:qualitative_mv}%
\end{figure}

The quantitative findings of Table~\ref{table:ehf_ablative} are
reflected in qualitative results.
In Figure~\ref{fig:qualitative}, we compare our final results with the initial baseline
that regresses all \smplx parameters directly from a low-resolution image without any
attention (first row in Tab. \ref{table:ehf_ablative}).
We observe that our body-attention mechanism gives a clear improvement for the hand and the face area.
Figure~\ref{fig:qualitative_mv} contains \modelname reconstructions, seen from multiple views, where we
again see the higher level of detail offered by our method.
For more qualitative results, see \supmat

\section{Conclusion}
\label{sec:conclusion}

In this paper, we present a regression approach for holistic expressive body reconstruction.
Considering the different scale of the individual parts and the limited training data,
we identify that the naive approach of regressing a holistic
reconstruction from a low-resolution body image misses fine details in
the hands and face.
To improve our regression approach, we investigate a body-driven attention
scheme.
This results in consistently better reconstructions.
Although the pure optimization-based
approach~\cite{Pavlakos_2019_CVPR}  recovers the finer details, it is
too slow to be practical.
\modelname provides competitive results, while more than two orders of magnitude faster than~\cite{Pavlakos_2019_CVPR}.
Eventually the two approaches could be combined effectively, as in \cite{Kolotouros_2019_ICCV}.
Considering the level of the accuracy and the speed of our approach,
we believe it should be a valuable tool and enable many applications
that require expressive human pose information. %
Future work will
extend the inference to multiple
humans~\cite{Jiang_2020_CVPR,zanfir_2018_cvpr,zanfir_nips_2018} and
video sequences ~\cite{kanazawa_2019_cvpr,Kocabas_2020_CVPR}.
The rich body representation will also accelerate research on
human-scene \cite{PROX:2019,savva2016pigraphs} interaction,
human-object \cite{Laptvev_CVPR_2019_forces,GRAB:2020} interaction,
and person-person
interaction~\cite{Fieraru_2020_CVPR,li_2020_interactingHumans}.
We also plan to improve body shape estimation and the pixel-level alignment to the image.

{
\smallskip
\noindent
\footnotesize
\textbf{\emph{Acknowledgements:}}
We thank Haiwen Feng for the \flame fits,
Nikos Kolotouros, Muhammed Kocabas and Nikos Athanasiou for helpful discussions,
Sai Kumar Dwivedi and Lea M{\"u}ller for proofreading,
Mason Landry and Valerie Callaghan for video voiceovers.
This research was partially supported by the Max Planck ETH Center for Learning Systems.
\textbf{\emph{Disclaimer:}}
MJB has received research gift funds from Intel, Nvidia, Adobe, Facebook, and Amazon.
While MJB is a part-time employee of Amazon, his research was performed solely at, and funded solely by, MPI.
MJB has financial interests in Amazon and Meshcapade GmbH.
}

\clearpage
\bibliographystyle{splncs04}
\bibliography{bibliography/final_bibliography}

% \includepdf[pages=-]{sup_mat.pdf}


\section*{Supplementary material}

# Monocular Expressive Body Regression through Body-Driven Attention
## *Supplemental Material*

Vasileios Choutas[1,2], Georgios Pavlakos[3], Timo Bolkart[1],
Dimitrios Tzionas[1], and Michael J. Black[1]

[1] Max Planck Institute for Intelligent Systems, Tübingen, Germany
[2] Max Planck ETH Center for Learning Systems
[3] University of Pennsylvania, Philadelphia, USA
{vchoutas, gpavlakos, tbolkart, dtzionas, black}@tuebingen.mpg.de

# https://expose.is.tue.mpg.de

Here we provide addition details and visualizations of our results. Please also see the narrated **video** on our website for a summary of the method and results.

## 1   Training details

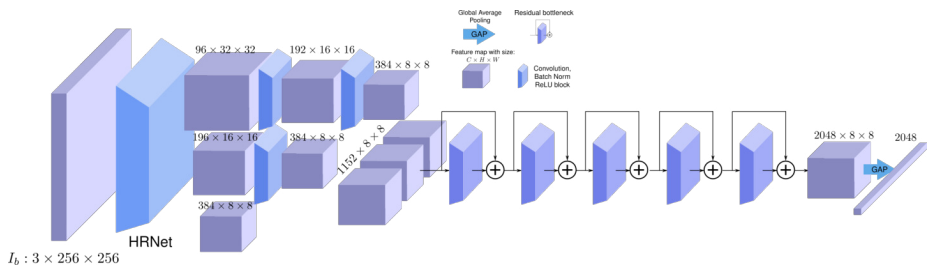

Fig. 1: Structure of the feature extractor used by the body prediction network. The image $I_b$ is fed to HRNet [9] to extract multi-scale feature maps. These are then processed by extra convolutional blocks and downsampled to the same spatial resolution. All feature maps are subsequently concatenated and fed to 5 residual blocks [1], followed by a global average pooling operation that produces the final feature vector $\mathcal{F}_b$.

**Architecture**: The features $\mathcal{F}_b$ are extracted from the body image $I_b$ using the architecture of Figure 1. The parameters $\Theta = \left\{\boldsymbol{\beta}, \boldsymbol{\theta}, \boldsymbol{\psi}, s, \bar{t}\right\}$ are predicted by feeding the features $\mathcal{F}_b$ and the mean parameters $\bar{\Theta}$ to an iterative regression network, whose structure follows [2]. The composition of the feature extraction network of Figure 1 and the iterative regressor forms the body network $g$.

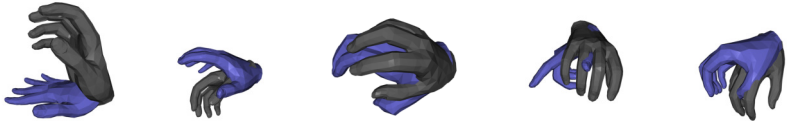

Fig. 2: **Illustrative examples.** The default global rotation of the hand is replaced by a random rotation with angle $r_{\text{global}} \sim \mathcal{U}(r_{\text{min}}, r_{\text{max}})$ around the ground truth axis of rotation given by the training data. We selected a range $(r_{\text{min}}, r_{\text{max}})_{\text{hand}} = (-90, 90)$ degrees. Blue is the ground-truth mesh used as a target for training, while gray is the starting point of the iterative hand regressor with a perturbed global rotation.

**Training**: We pre-train the body network until validation performance on 3DPW [6] saturates, using Adam [4], with batch size 48. The hand and head sub-networks are pre-trained as well on the FreiHand [10] and FFHQ [3] data, again with Adam [4] and a batch size of 64. Once validation performance saturates, we freeze the body network and fine-tune the hand and head sub-networks with all available training data to produce ExPose. The exact hyper-parameters will be included in the released code. The entire pipeline is implemented in PyTorch.

## 2 Data augmentation

For hand and face-only data, shape and pose regression is done following the iterative scheme of [2], which computes offsets from a set of mean parameters. When we have access to full body information, we wish to condition the part specific sub-networks on the output of the body network. However, naively adding this conditioning is not enough, as this creates a domain gap between hands and face-only images and those coming from the body attention mechanism. To bridge this, we augment the training data by modifying the initial mean point to some random point. In this way, the part sub-network will be forced to learn to predict the correct offsets, no matter the initial point, that lead to the pose and shape that matches the image. As described in the main text, we randomly perturb the global rotation of the hand and face data around the ground-truth axis of rotation, as illustrated in Figures 2 and 3 respectively. We also modify the shape of the hand and the face by randomly sampling from normal distributions over the hand and face shape parameters, as illustrated in Figures 6 and 5 respectively. For the face-only data, we also augment the rotation of the jaw, by replacing the default value with a random rotation around the x-axis, seen in Figure 4. Finally, we replace the default mean expression with a sample drawn from a standard normal distribution, as seen in Figure 7.

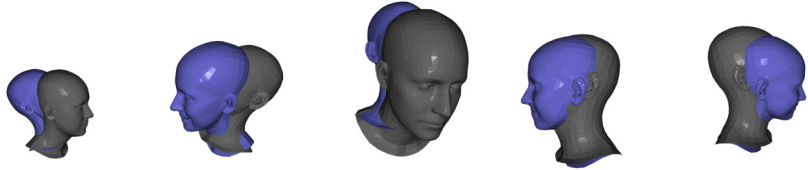

Fig. 3: The default global rotation of the head is replaced by a random rotation with angle $r_{\text{global}} \sim \mathcal{U}(r_{\min}, r_{\max})$ around the ground truth axis of rotation given by the training data. We selected a range $(r_{\min}, r_{\max})_{\text{head}} = (-45, 45)$ degrees. Blue is the ground-truth mesh used as a target for training, while gray is the starting point of the iterative face regressor with a perturbed global rotation.

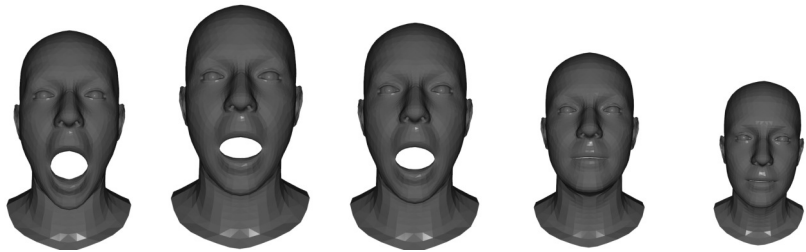

Fig. 4: The default rotation of the jaw, which corresponds to a closed mouth, is replaced by a random rotation around the x-axis. The angle of rotation is sampled randomly from the uniform distribution $r_{\text{jaw}} \sim \mathcal{U}(0, 45)$.

## 3   Converting SMPL to SMPL-X

There exist a wide variety of SMPL annotations for training 3D body pose and shape estimation methods. It is therefore important to create an automated method to convert them to the corresponding SMPL-X parameters, to use them as training data. To achieve this, we leverage the relation between SMPL and SMPL-X to build a correspondence map between the two models. SMPL and SMPL-X are articulated models of the human body that produce 3D triangle meshes:

$$M_{\text{SMPL}} = (V_{\text{SMPL}}, F_{\text{SMPL}}), \ V_{\text{SMPL}} \in \mathbb{R}^{6890 \times 3}, F_{\text{SMPL}} \in \mathbb{N}^{13776 \times 3} \tag{1}$$

$$M_{\text{SMPL-X}} = (V_{\text{SMPL-X}}, F_{\text{SMPL-X}}), \ V_{\text{SMPL-X}} \in \mathbb{R}^{10475 \times 3}, F_{\text{SMPL-X}} \in \mathbb{N}^{20908 \times 3} \tag{2}$$

We start by registering the SMPL template mesh to the SMPL-X template. Given the registered meshes, we compute for each SMPL-X vertex $\boldsymbol{v}_i$ its nearest point $\boldsymbol{p}_i$ on the SMPL mesh and store the index of the nearest SMPL triangle $t_i$, its vertex indices $\boldsymbol{f_i} = [f_0^i, f_1^i, f_2^i]$ and the barycentric coordinates $[\alpha_i, \beta_i, \gamma_i]$ of $\boldsymbol{p}_i$ with respect to triangle $t_i$. We also store a binary mask $m_i$ for each vertex

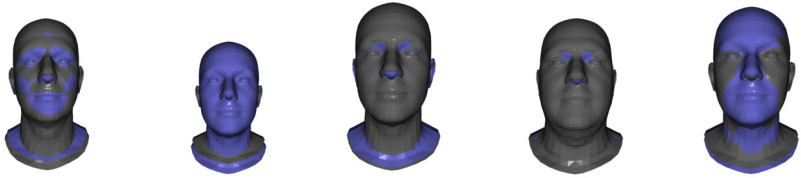

Fig. 5: The default mean shape of the head is replaced with a random vector $\boldsymbol{\beta} \sim \mathcal{N}\left(\vec{0}, I\right), I \in \mathbb{R}^{10 \times 10}$ The blue mesh represents the mean shape, while the gray mesh has a random shape drawn from the above distribution.

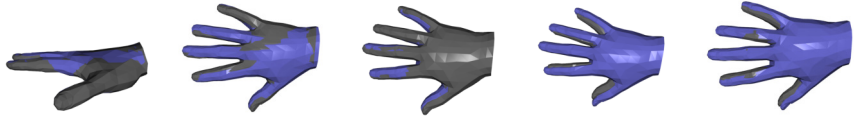

Fig. 6: The default mean shape of the hand is replaced with a random vector $\boldsymbol{\beta} \sim \mathcal{N}\left(\vec{0}, I\right), I \in \mathbb{R}^{10 \times 10}$ The blue mesh represents the mean shape, while the gray mesh has a random shape drawn from the above distribution.

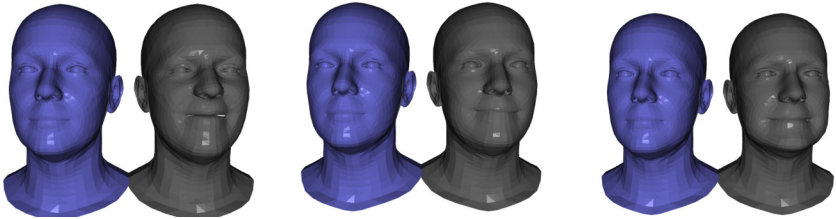

Fig. 7: The default neutral expression of the head is replaced with a random vector $\boldsymbol{\psi} \sim \mathcal{N}\left(\vec{0}, I\right), I \in \mathbb{R}^{10 \times 10}$. The blue mesh represents the neutral expression, while the gray mesh has a random expression drawn from the above distribution.

that is used to mask invalid correspondences between the two models, such as the eyes, inner lip region, etc.

Given a posed SMPL mesh $M' = (V', F')$, e.g. one sample from the fit data of SPIN [5], we build a mesh $\hat{M}$ in SMPL-X topology. Vertex $\hat{v}_i$ of the mesh $\hat{M}$ is computed as:

$$\hat{v}_i = \alpha_i v'_{f_0^i} + \beta_i v'_{f_1^i} + \gamma_i v'_{f_2^i} \tag{3}$$

where $v_{f_0^i}'$ is the SMPL vertex with index $f_0^i$. We now have a mesh in SMPL-X topology, which we will use to find the corresponding pose $\boldsymbol{\theta}$, shape $\boldsymbol{\beta}$, expression $\psi$ and translation $\boldsymbol{t}$ parameters. Let $v_i$ be the i-th vertex returned by posing SMPL-X using the current values of the parameters $(\boldsymbol{\theta}, \boldsymbol{\beta}, \boldsymbol{\psi}, \boldsymbol{t})$. We start by

optimizing only over the pose $\boldsymbol{\theta}$ using the following loss:

$$L_1(\boldsymbol{\theta}) = \sum_{(i,j)\in\mathcal{E}} m_i m_j \left\| (v_i - v_j) - (\hat{v}_i - \hat{v}_j) \right\|_2^2 \tag{4}$$

where $\mathcal{E}$ is the set of 3D edges of the SMPL-X mesh. We use the binary masks $m_i, m_j$ to compute the loss only on valid vertices. For the second stage, we optimize the translation vector $\boldsymbol{t}$ using a vertex-to-vertex loss:

$$L_2(\boldsymbol{t}) = \sum_i m_i \left\| v_i - \hat{v}_i \right\|_2^2 \tag{5}$$

By this point, we have rigidly aligned the two meshes and matched the articulation of the original SMPL mesh. All that remains is to also match the shape, to get the best possible fit. The final step is to optimize over all parameters $(\boldsymbol{\theta}, \boldsymbol{\beta}, \boldsymbol{\psi}, \boldsymbol{t})$ using again a vertex-to-vertex loss:

$$L_3((\boldsymbol{\theta}, \boldsymbol{\beta}, \boldsymbol{\psi}, \boldsymbol{t})) = \sum_i m_i \left\| v_i - \hat{v}_i \right\|_2^2 \tag{6}$$

We use a Trust Region Newton Conjugate Gradient optimizer [7] to search for minimize the objectives. The implementation for the transfer process can be in found on our website: https://expose.is.tue.mpg.de.

## 4  SMPLify-X qualitative comparison

As shown in Table 3 of the main manuscript, ExPose is almost $200\times$ times faster compared to SMPLify-X [8], and provides qualitatively similar results to the latter, as seen in Figures 8a and 8b. Although the accuracy of ExPose is slightly lower than SMPLify-X, it can provide a better initialization to the latter, helping it overcome failures of its initialization heuristic and of the keypoint detector. Potentially, this could be done in a loop, similar to [5] to continuously improve the performance of ExPose using mode in-the-wild data.

## 5  In-the-wild qualitative results

A qualitative comparison of our method with the state-of-the-art SMPL regression methods shows the increase in expressivity offered by ExPose; see Figures 10 to 20. Figure 9 compares the output of the naive regression approach with the body-driven attention mechanism of ExPose. Finally, Figures 21 to 25 contain visualizations of ExPose predictions from multiple views.

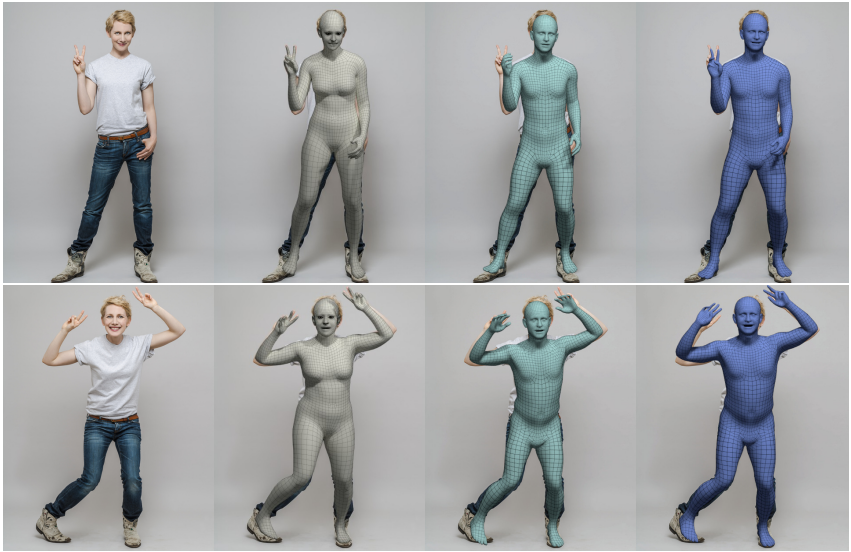

(a) 1. The input image, 2. SMPLify-X (known gender), 3. naive regression from a single body image fails to capture detailed finger articulation and facial expressions, 4. ExPose is able to recover these details, thanks to its attention mechanism, and produces results of similar quality as SMPLify-X, while being 200 times faster.

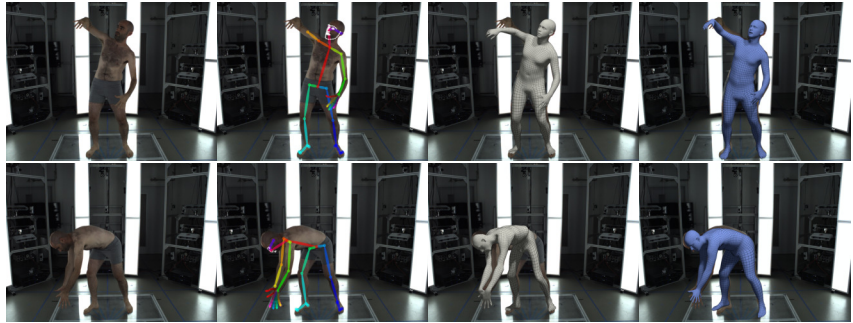

(b) 1. The input image, 2. OpenPose detections, 3. SMPLify-X fitting, with the neutral model and default focal length, 4. ExPose. When 2D keypoint detections are missing or wrong, optimization based methods, such as SMPLify-X are unable to avoid implausible poses. Furthermore, they heavily depend on their initialization and can produce unnatural poses and shapes, when their initialization heuristic fails. Regression methods, such as ExPose, avoid these problems and can provide better initialization points, closer to the actual solution, and accelerate convergence.

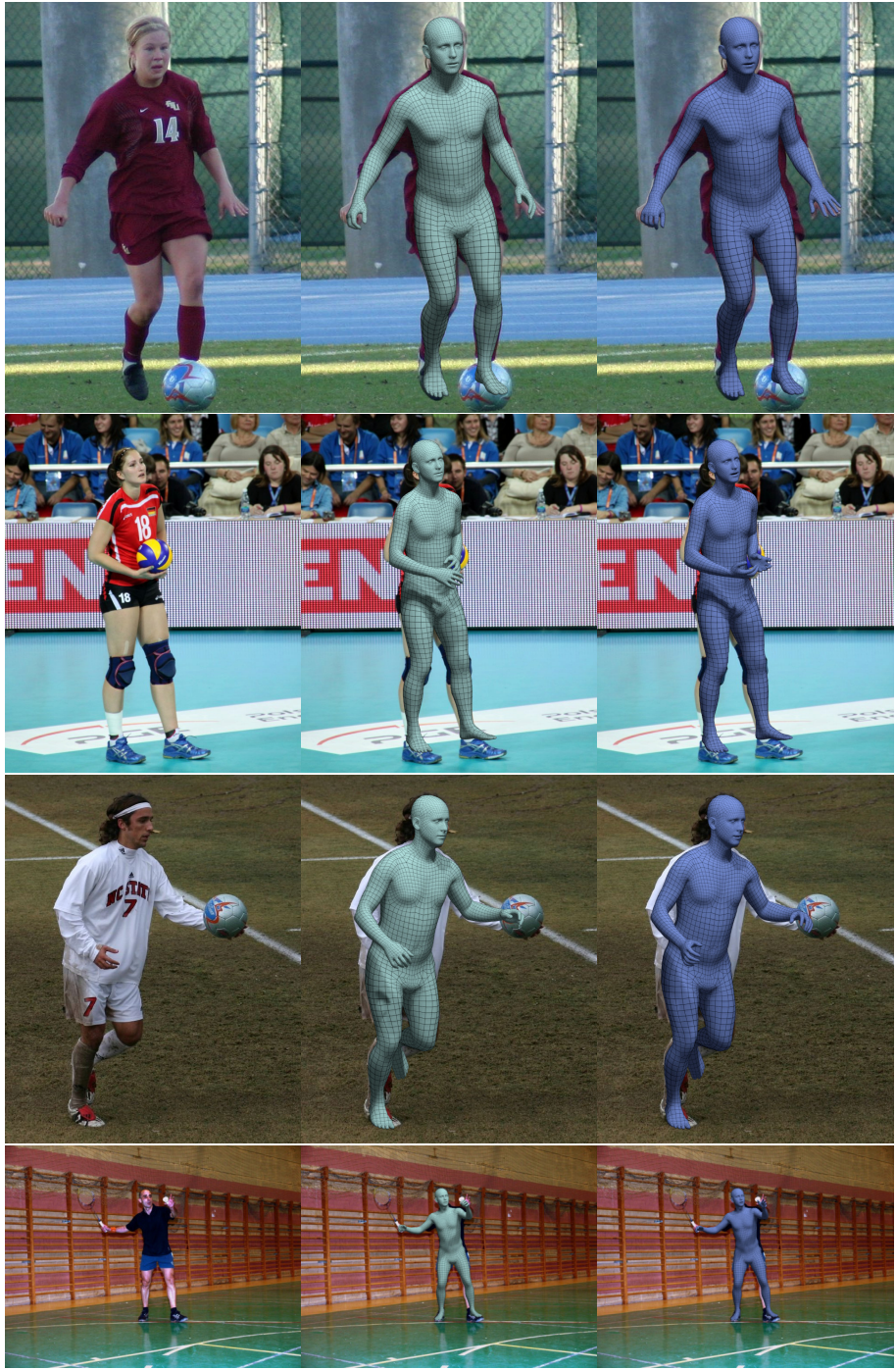

Fig. 9: *Left:* The input image. *Middle:* Naive regression from a body crop. *Right:* ExPose. The attention mechanism helps capture detailed hand articulation and facial expression.

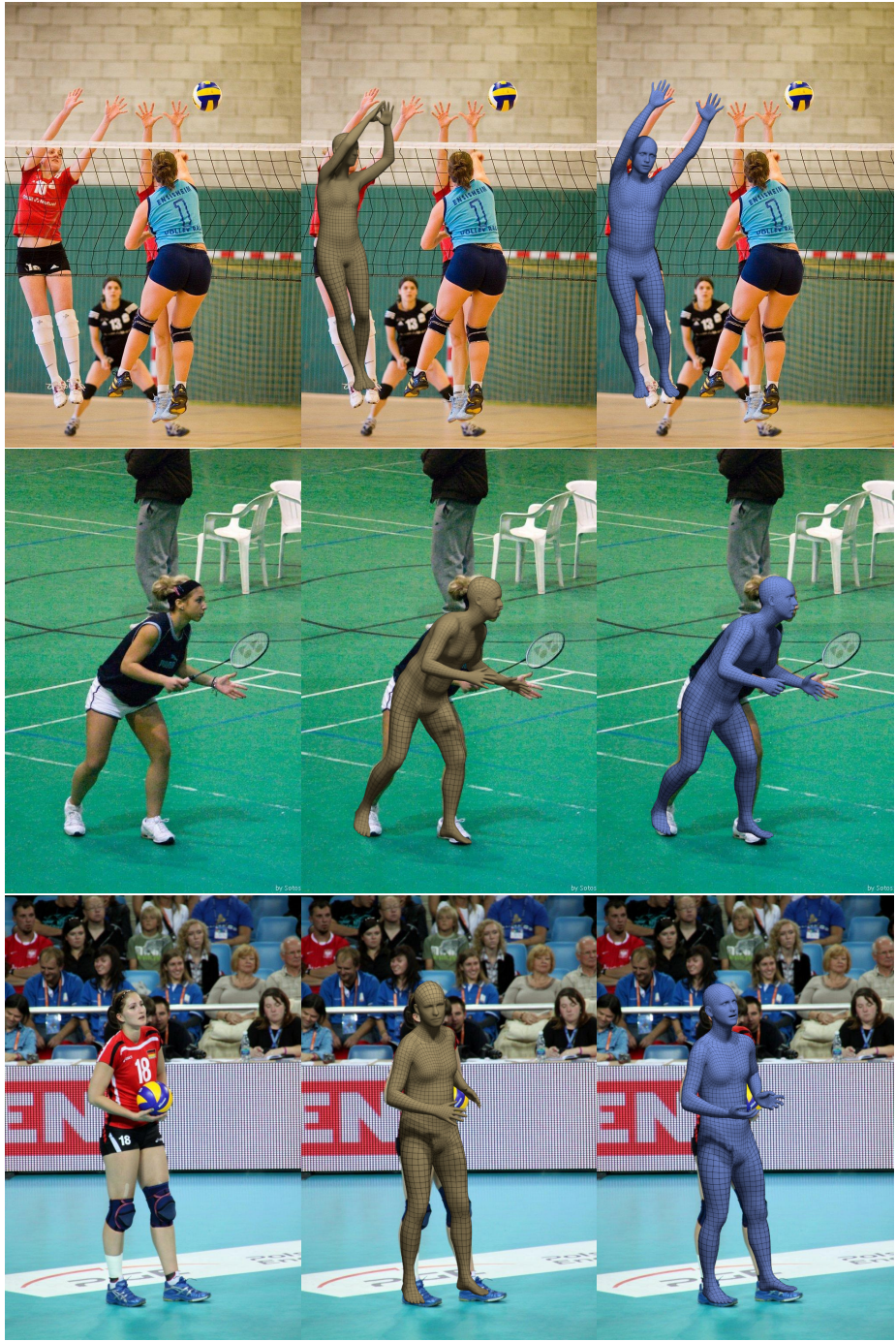

Fig. 10: Comparison of ExPose with the state-of-the-art body regression method:
1. RGB image, 2. SPIN [5], 3. ExPose. Our proposed method produces 3D body
pose and shape results on par with SPIN [5] and captures more details for the
hands and face.

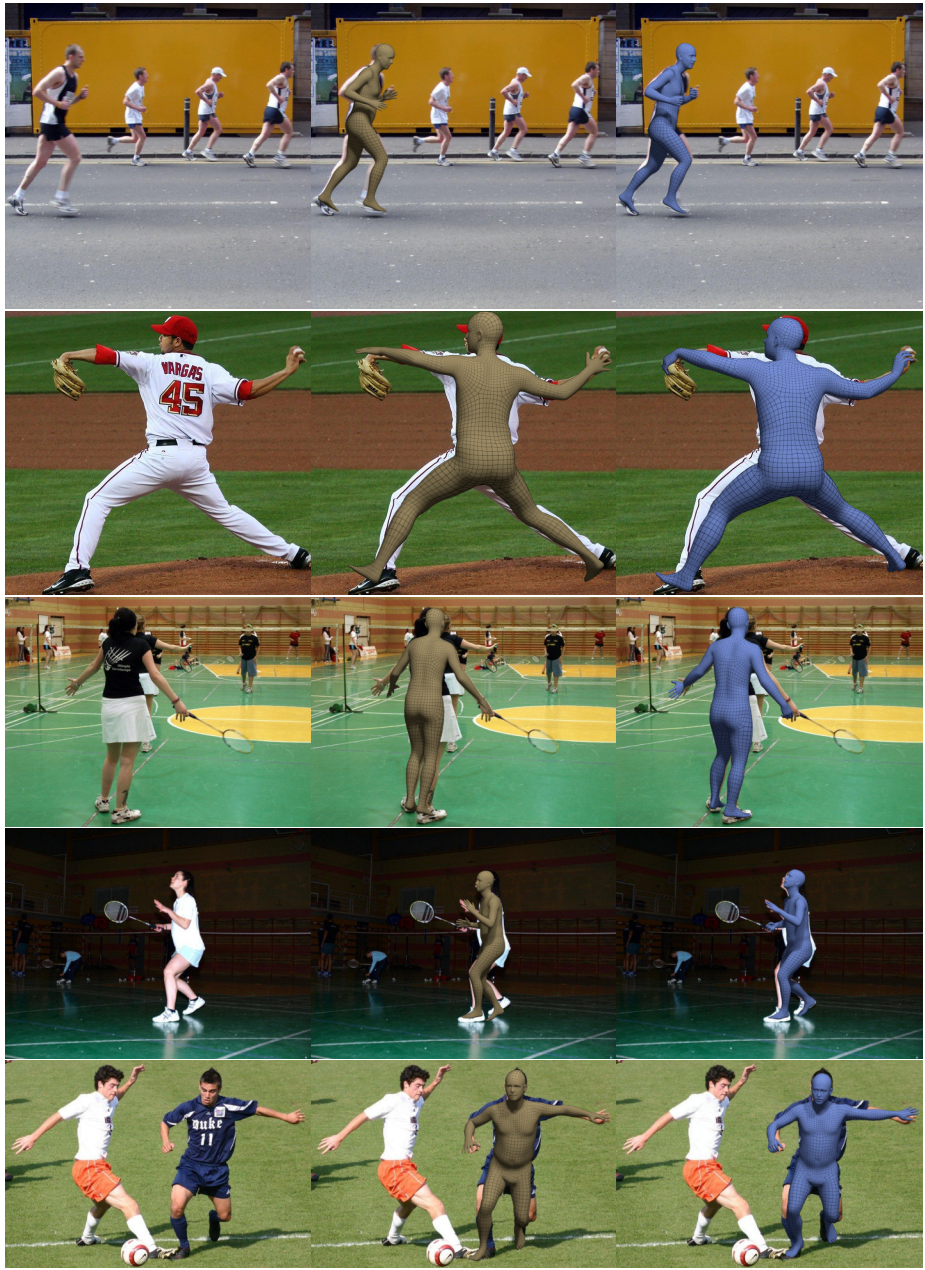

Fig. 11: Comparison of ExPose with the state-of-the-art body regression method: 1. RGB image, 2. SPIN [5], 3. ExPose. Our proposed method produces 3D body pose and shape results on par with SPIN [5] and captures more details for the hands and face.

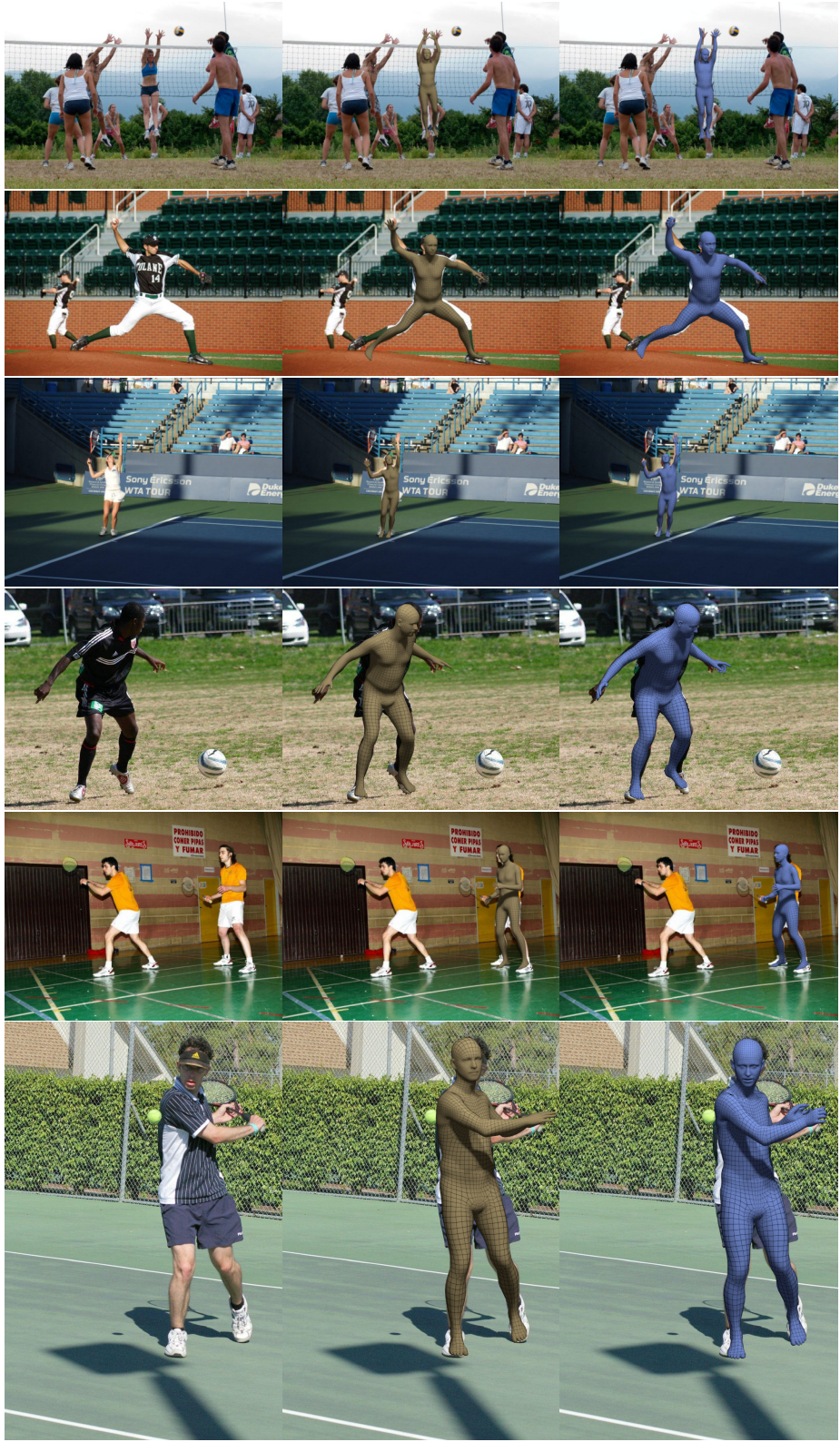

Fig. 12: Comparison of ExPose with the state-of-the-art body regression method: 1. RGB image, 2. SPIN [5], 3. ExPose. Our proposed method produces 3D body pose and shape results on par with SPIN [5] and captures more details for the hands and face.

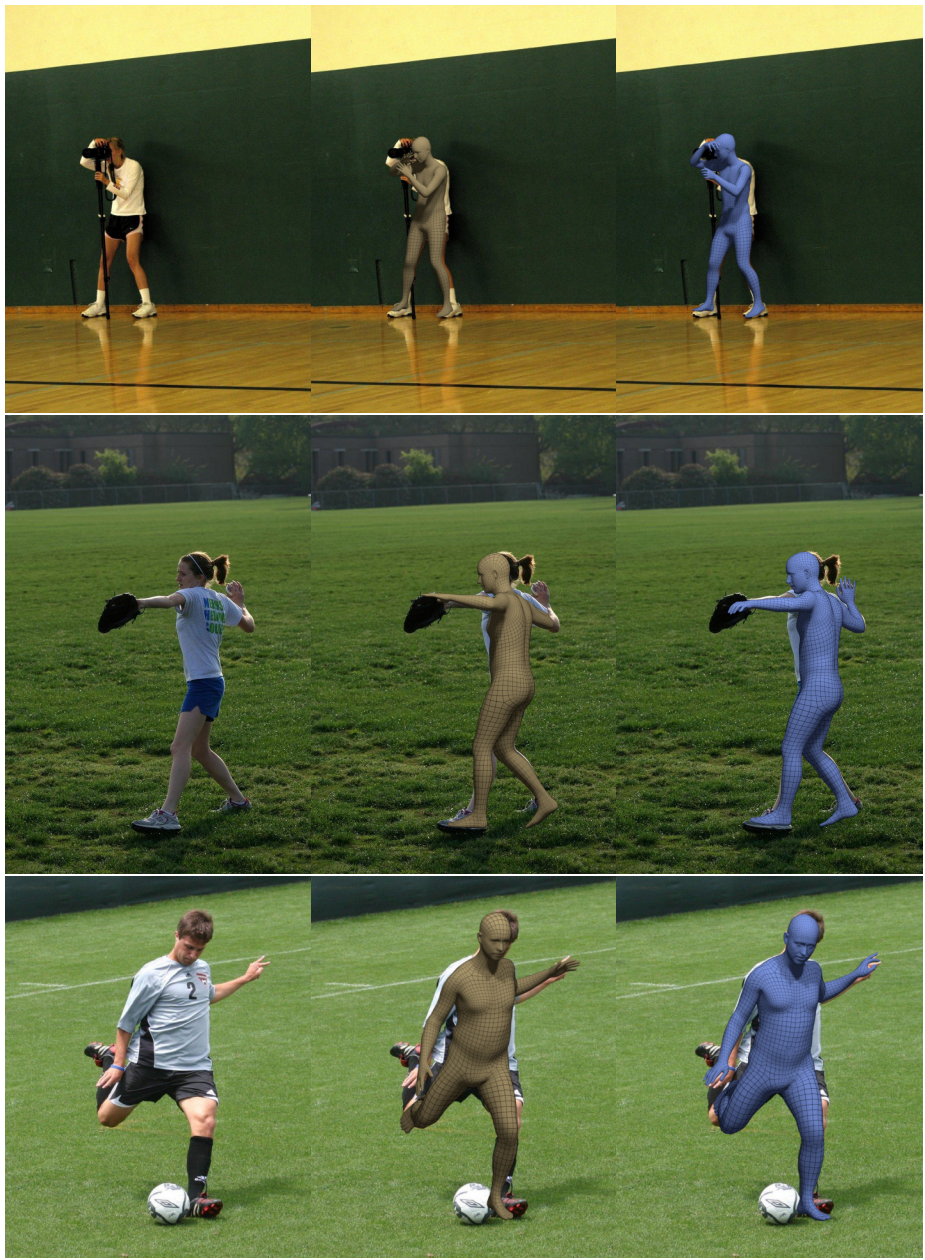

Fig. 13: Comparison of ExPose with the state-of-the-art body regression method: 1. RGB image, 2. SPIN [5], 3. ExPose. Our proposed method produces 3D body pose and shape results on par with SPIN [5] and captures more details for the hands and face.

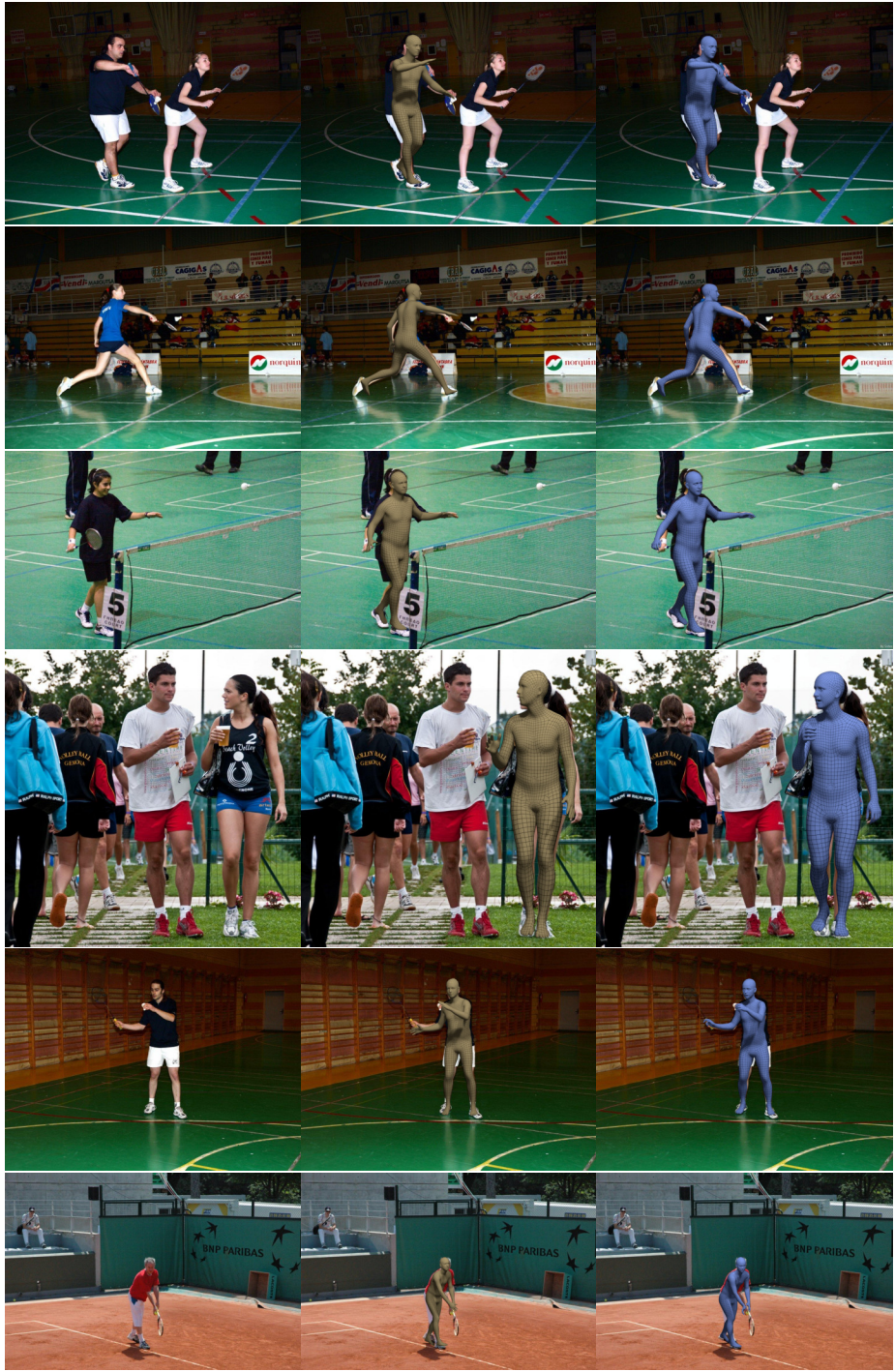

Fig. 14: Comparison of ExPose with the state-of-the-art body regression method:
1. RGB image, 2. SPIN [5], 3. ExPose. Our proposed method produces 3D body
pose and shape results on par with SPIN [5] and captures more details for the
hands and face.

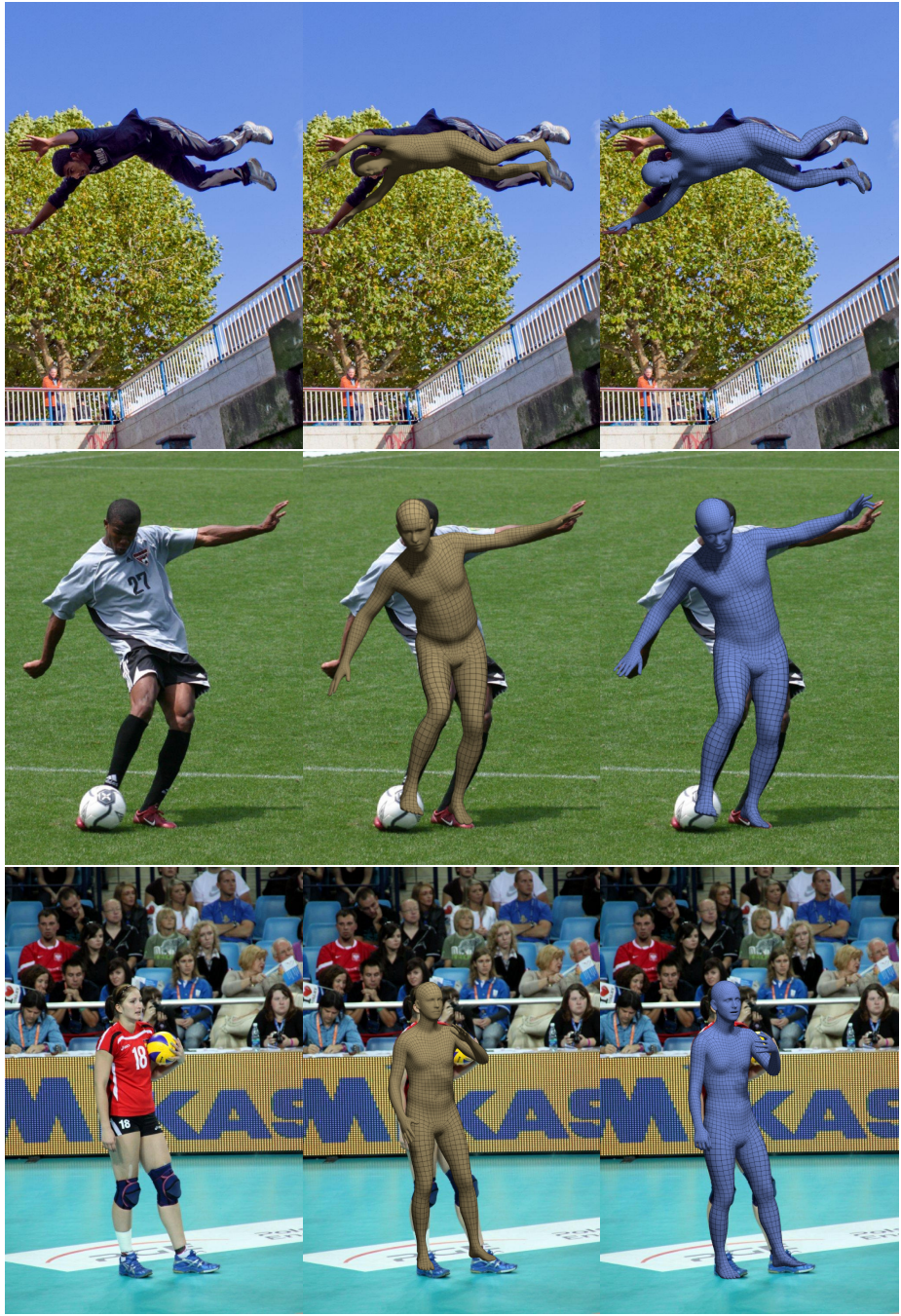

Fig. 15: Comparison of ExPose with the state-of-the-art body regression method: 1. RGB image, 2. SPIN [5], 3. ExPose. Our proposed method produces 3D body pose and shape results on par with SPIN [5] and captures more details for the hands and face.

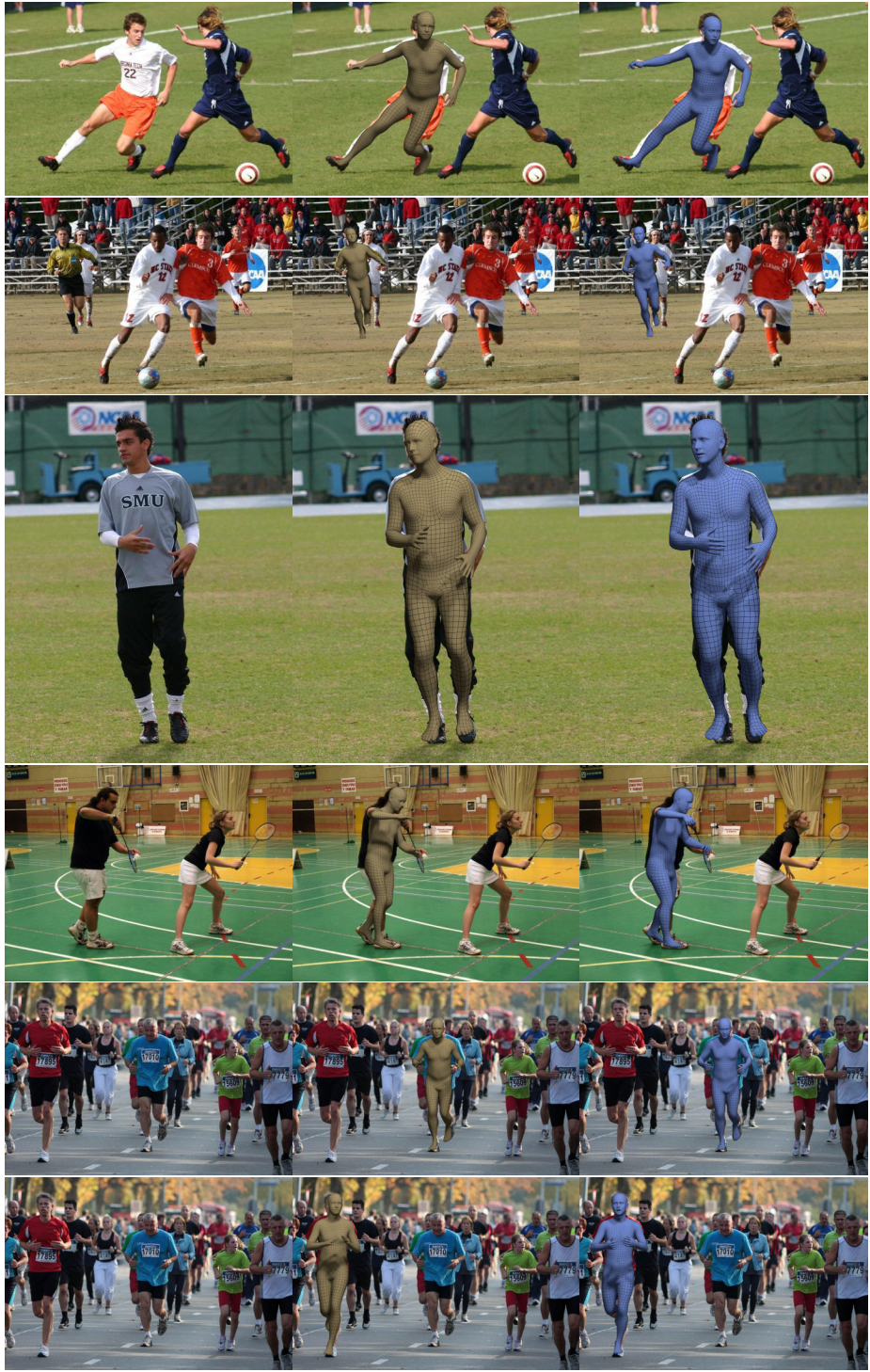

Fig. 16: Comparison of ExPose with the state-of-the-art body regression method:
1. RGB image, 2. SPIN [5], 3. ExPose. Our proposed method produces 3D body
pose and shape results on par with SPIN [5] and captures more details for the
hands and face.

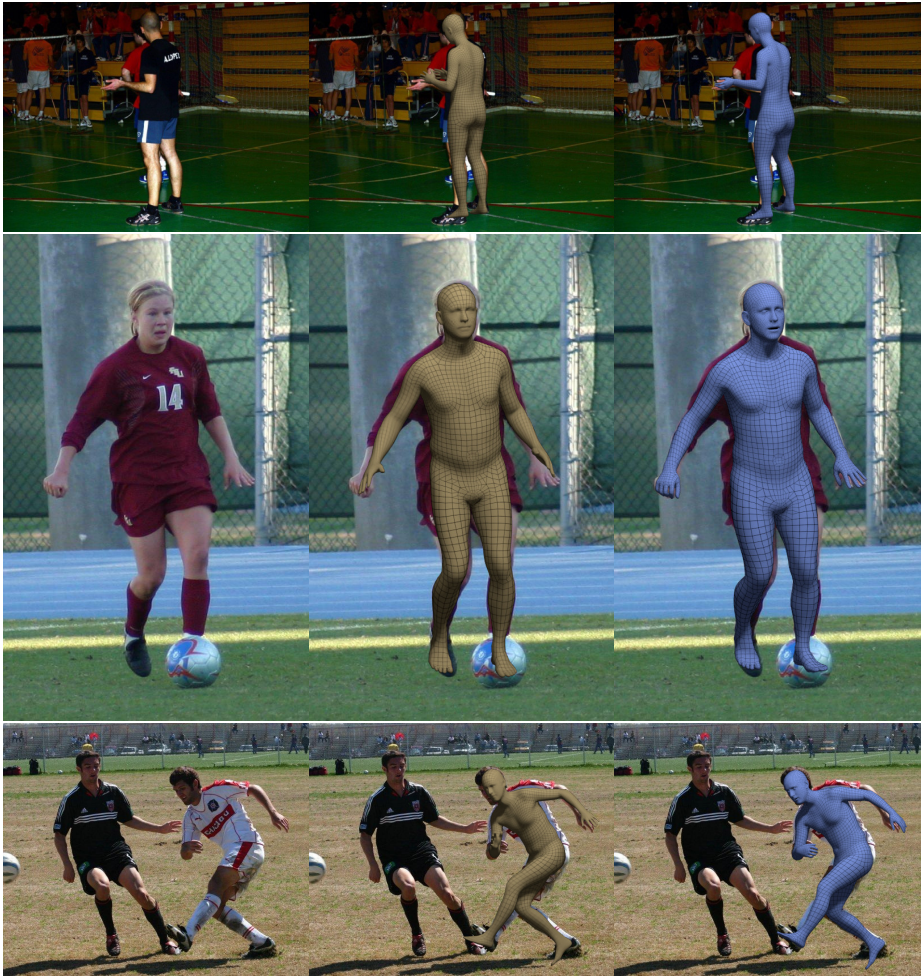

Fig. 17: Comparison of ExPose with the state-of-the-art body regression method: 1. RGB image, 2. SPIN [5], 3. ExPose. Our proposed method produces 3D body pose and shape results on par with SPIN [5] and captures more details for the hands and face.

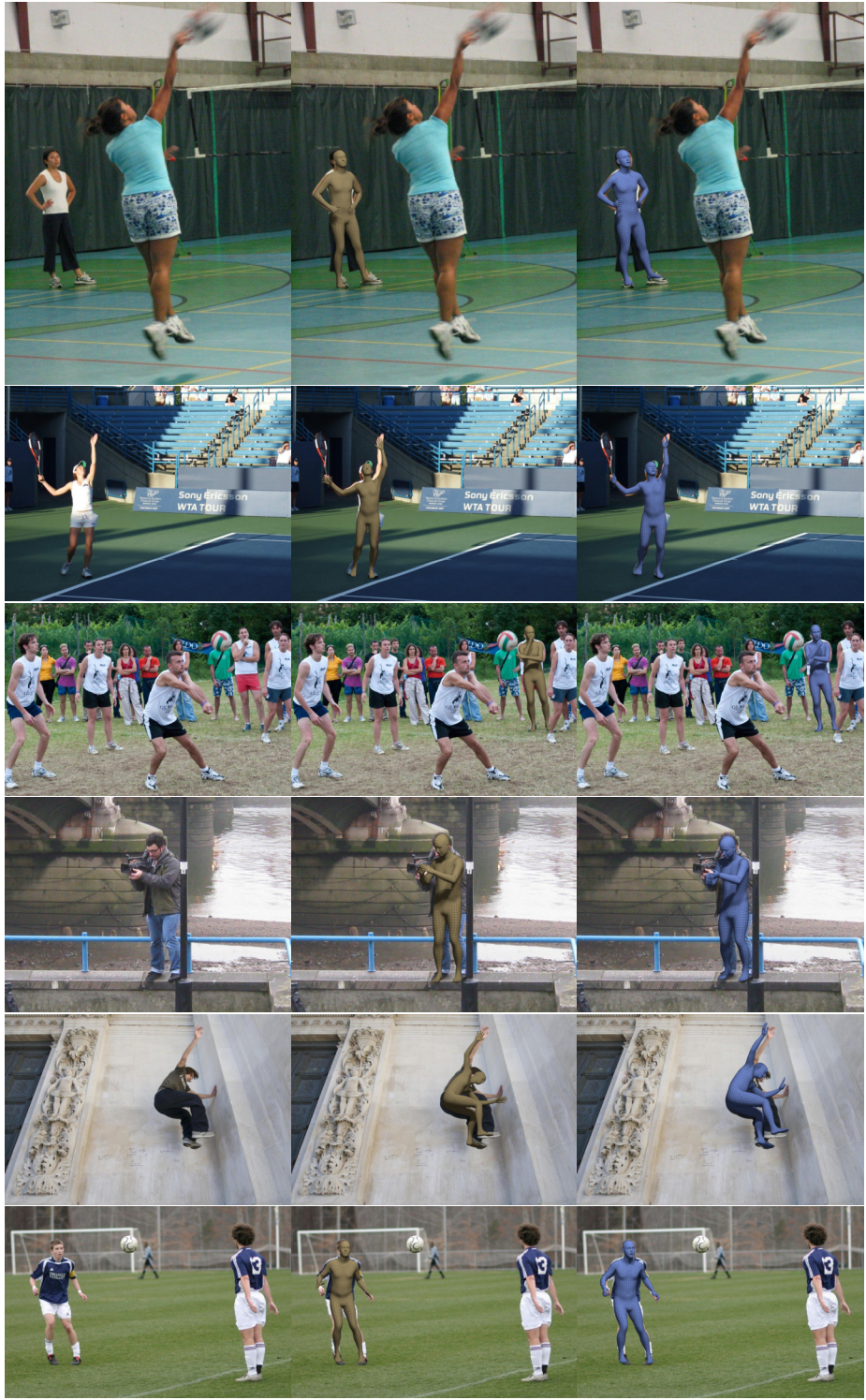

Fig. 18: Comparison of ExPose with the state-of-the-art body regression method: 1. RGB image, 2. SPIN [5], 3. ExPose. Our proposed method produces 3D body pose and shape results on par with SPIN [5] and captures more details for the hands and face.

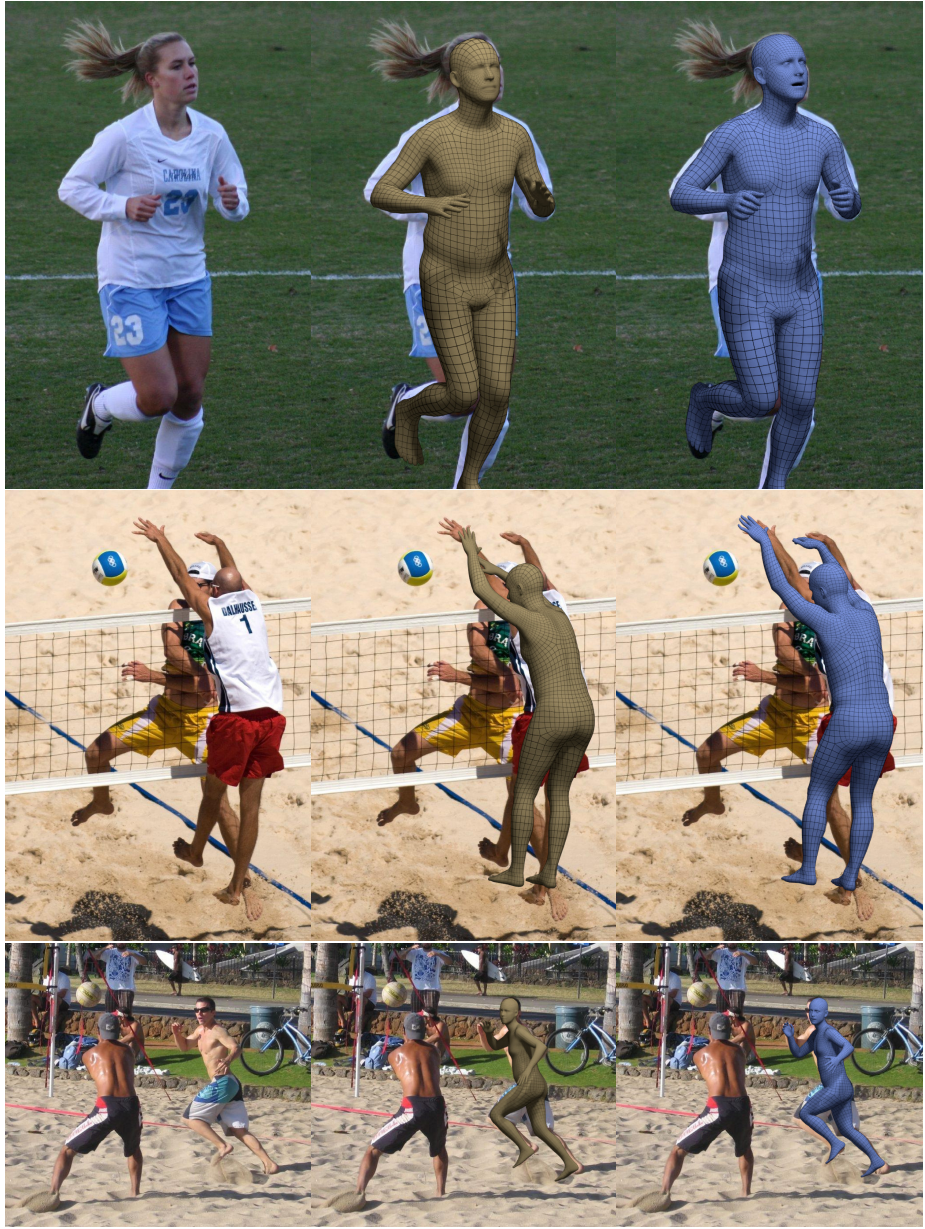

Fig. 19: Comparison of ExPose with the state-of-the-art body regression method: 1. RGB image, 2. SPIN [5], 3. ExPose. Our proposed method produces 3D body pose and shape results on par with SPIN [5] and captures more details for the hands and face.

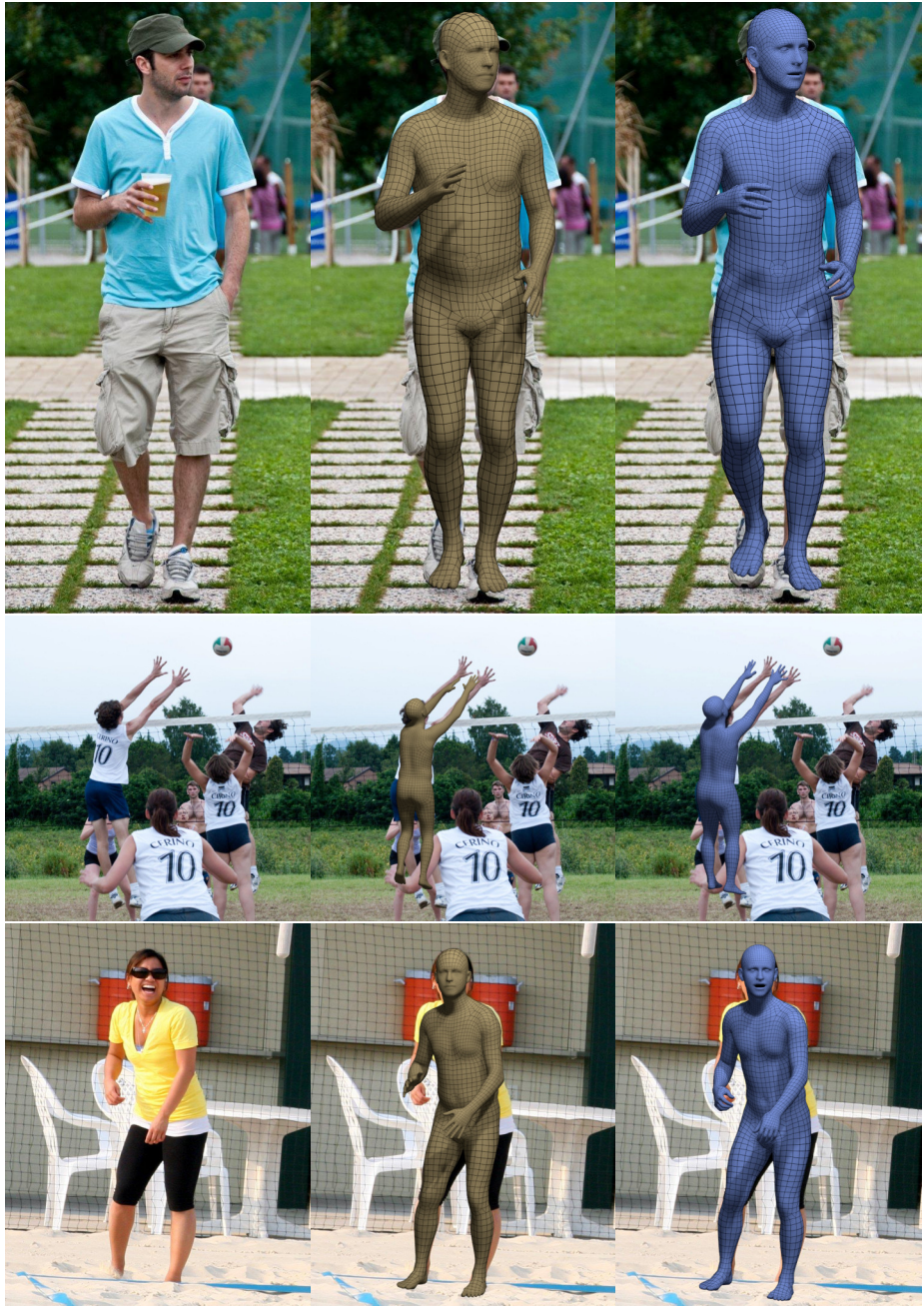

Fig. 20: Comparison of ExPose with the state-of-the-art body regression method: 1. RGB image, 2. SPIN [5], 3. ExPose. Our proposed method produces 3D body pose and shape results on par with SPIN [5] and captures more details for the hands and face.

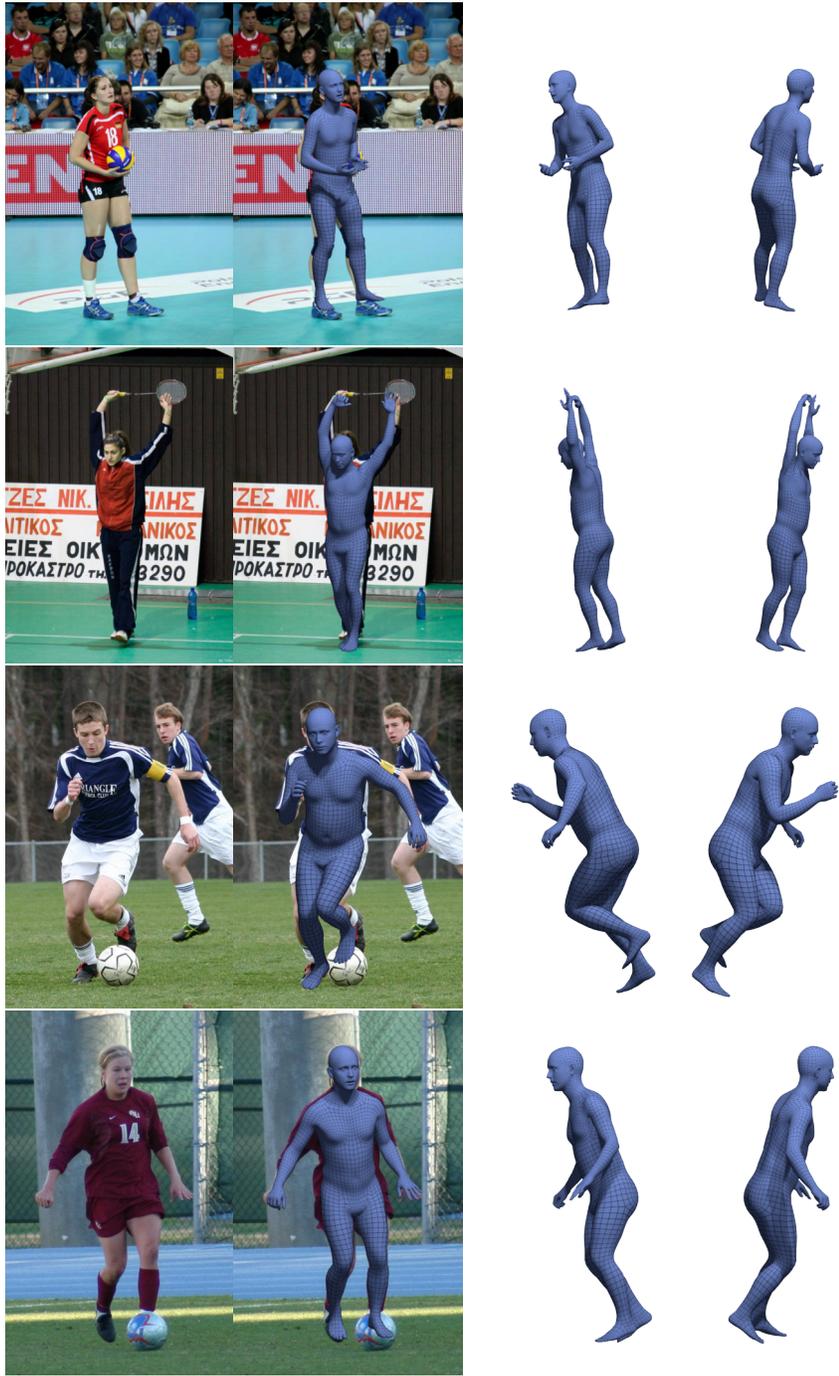

Fig. 21: ExPose results visualized from multiple views. 1. RGB image, 2. overlay, 3. , 4. rotations around the vertical axis

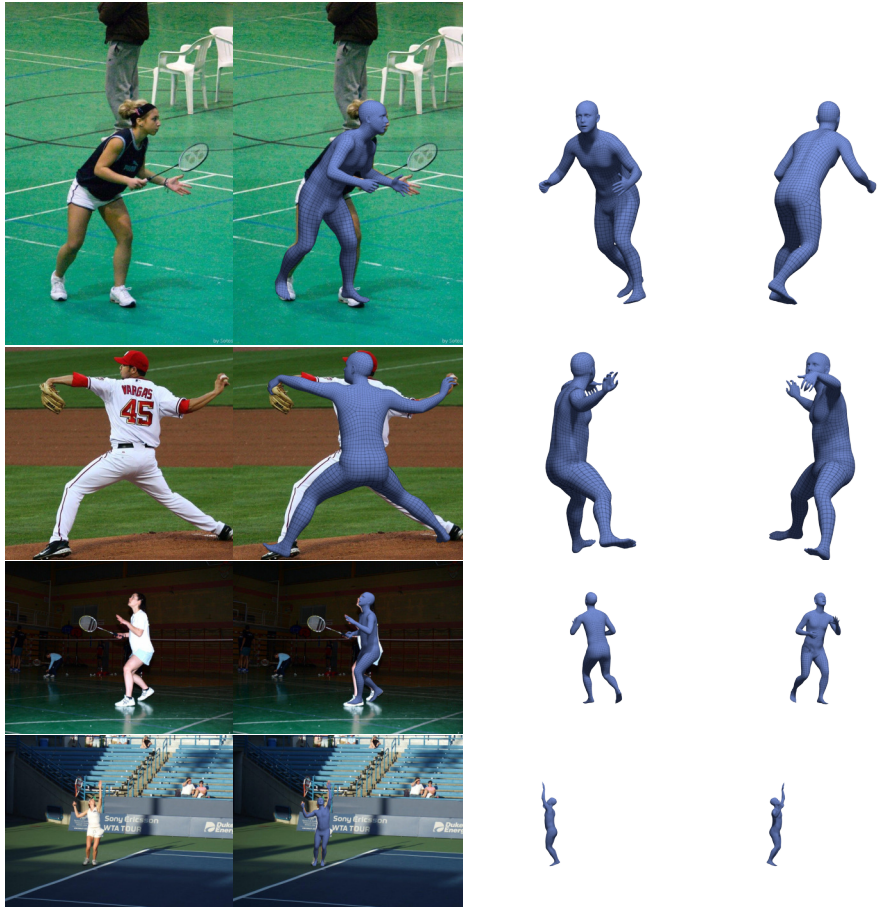

Fig. 22: ExPose results visualized from multiple views. 1. RGB image, 2. overlay, 3. , 4. rotations around the vertical axis

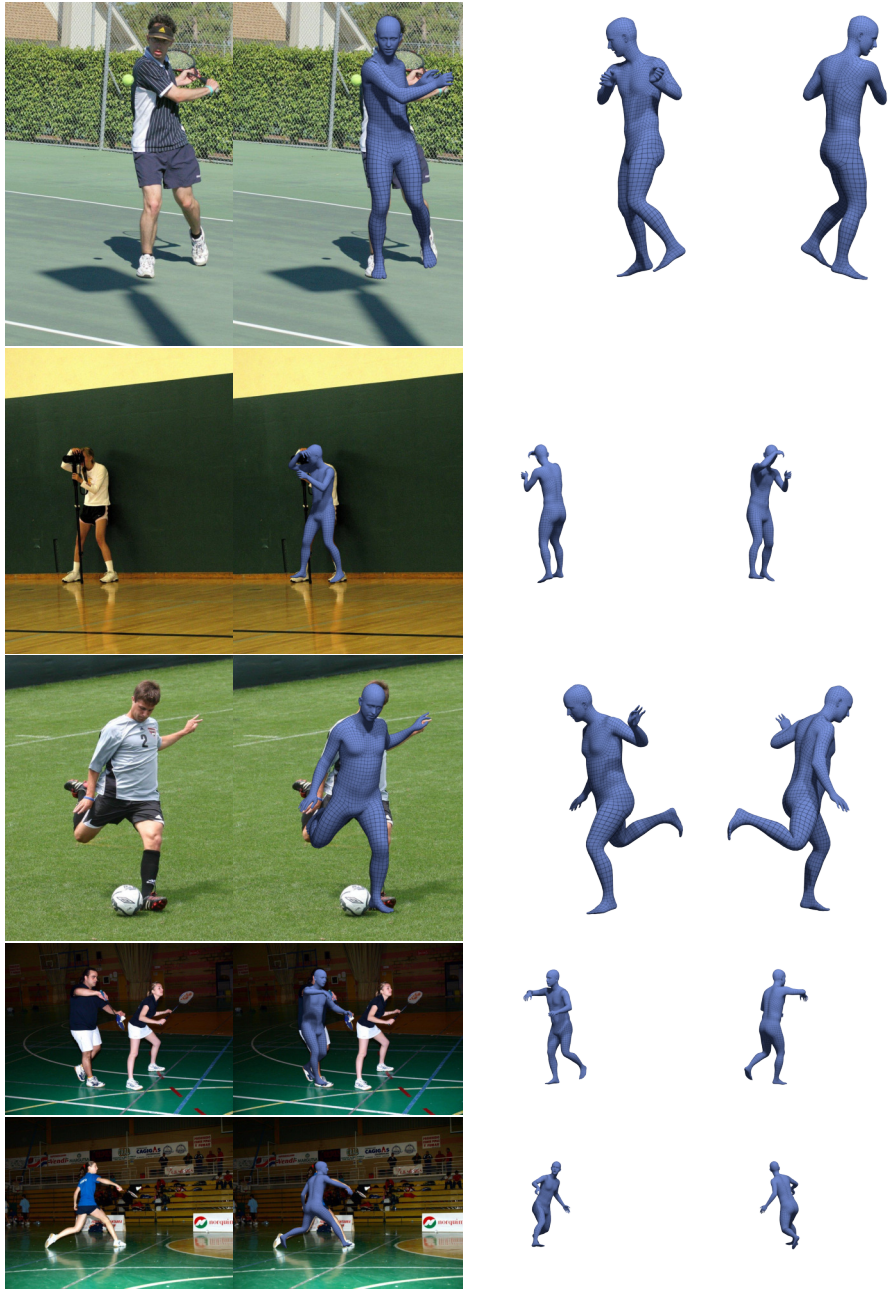

Fig. 23: ExPose results visualized from multiple views. 1. RGB image, 2. overlay, 3. , 4. rotations around the vertical axis

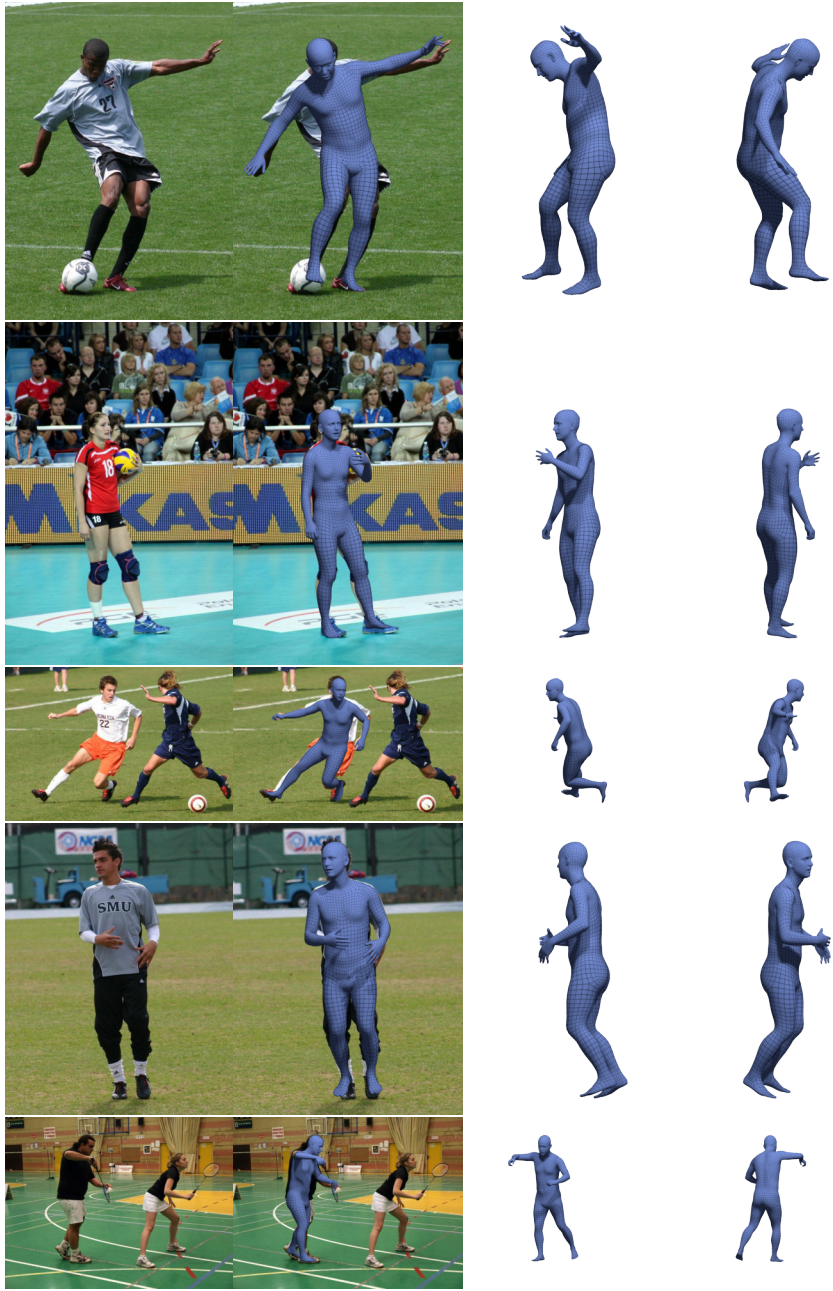

Fig. 24: ExPose results visualized from multiple views. 1. RGB image, 2. overlay, 3. , 4. rotations around the vertical axis

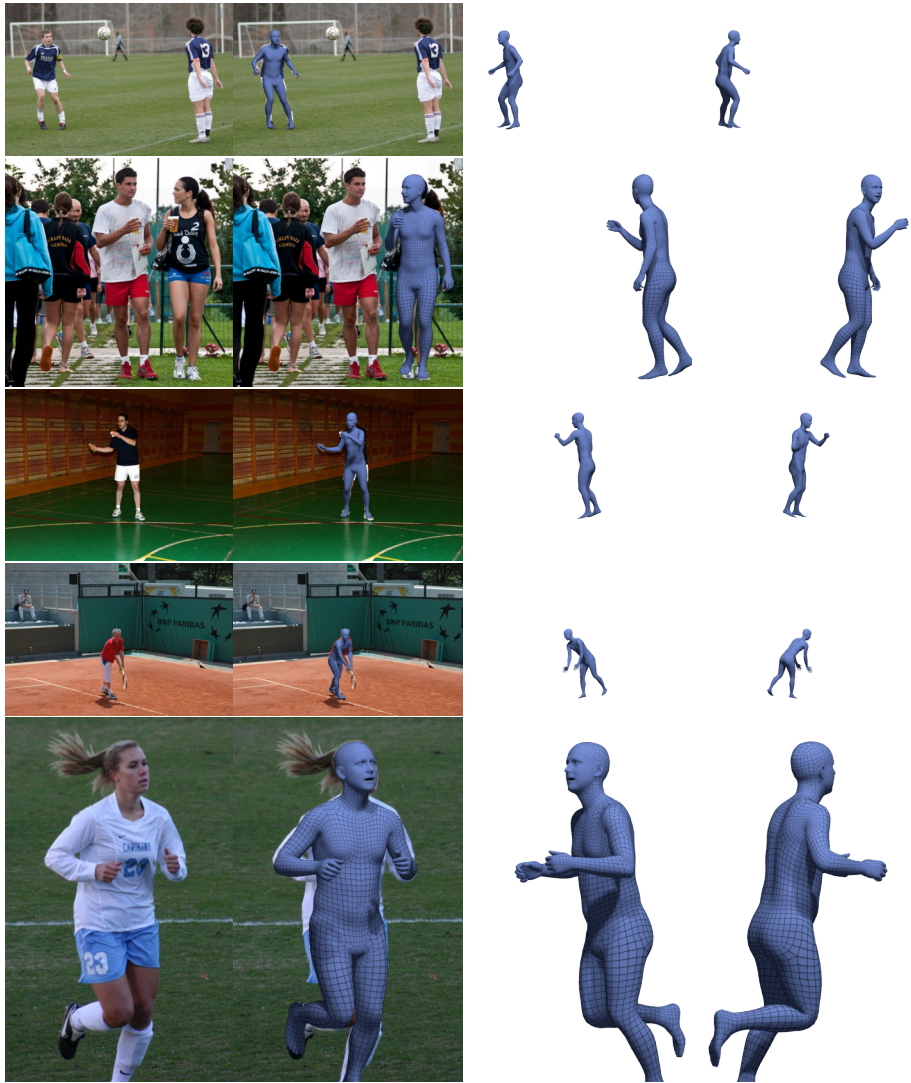

Fig. 25: ExPose results visualized from multiple views. 1. RGB image, 2. overlay, 3. , 4. rotations around the vertical axis



\end{document}